\documentclass[11pt]{article}

\usepackage[preprint]{acl}

\usepackage{microtype}

\usepackage[english,bidi=default]{babel}

\babelfont{rm}[
  BoldFont=TeXGyreTermesX-Bold.otf,
  ItalicFont=TeXGyreTermesX-Italic.otf,
  BoldItalicFont=TeXGyreTermesX-BoldItalic.otf
]{TeXGyreTermesX-Regular.otf}

\babelprovide[import]{arabic}

\babelfont[*arabic]{rm}[
  BoldFont=Amiri-Bold.ttf,
  ItalicFont=Amiri-Italic.ttf,
  BoldItalicFont=Amiri-BoldItalic.ttf
]{Amiri-Regular.ttf}

\usepackage{amsmath}
\usepackage{graphicx}
\usepackage{booktabs}
\usepackage{threeparttable}
\usepackage{multirow}
\usepackage{makecell}

\title{
Multilingual in Name Only? \\
Cultural and Linguistic Weaknesses of LLMs in Urdu
}

\author{
Farah Adeeba$^{1,2}$ \quad
Abdul Rafae Khan$^{3}$ \quad
Rajesh Bhatt$^{2}$ \quad
Hassan Sajjad$^{4}$ \\[4pt]
$^{1}$University of Konstanz, Germany \\
$^{2}$University of Massachusetts Amherst, USA \\
$^{3}$Malaysia School of Information Technology, Monash University Malaysia \\
$^{4}$Dalhousie University, Canada \\[4pt]
\texttt{farah.adeeba@uni-konstanz.de} \quad
\texttt{Abdul.Rafae@monash.edu} \\
\texttt{bhatt@umass.edu} \quad
\texttt{HSajjad@dal.ca}
}

\begin{document}

\maketitle

\begin{abstract}
Multilingual large language models (LLMs) are increasingly used for open-ended text generation, yet their behaviour in low-resource languages remains poorly understood. In this work, we question how correct and reliable is the generation of multilingual LLMs when used for the task of story generation. We consider Urdu language as a representative low-resource language. We generate \textbf{Urdu-Stories}, a corpus of 93 stories generated using three contemporary LLMs (GPT-5.1, Qwen-3-Max, DeepSeek-3.1). We manually annotate the errors present in them under a nine-label linguistic, semantic, and cultural taxonomy. Our notable findings suggest that LLMs often make basic errors of grammar and semantics. The stories lack coherence, have unnatural repetition and show pervasive cultural shallowness. We further show using few-shot prompting that the cultural and context errors largely remain unresolved. Our findings highlight the limitations of current LLMs as a reliable source of content generation and information retrieval for low-resource languages.  
\end{abstract}

\section{Introduction}
Large Language Models (LLMs) have demonstrated remarkable performance across a wide range of text generation tasks, yet this success remains concentrated predominantly in English and a few other handful of languages~\cite{joshi2020state,blasi2022systematic}. While many of these models are described as multilingual and are capable of generating text in numerous languages, their performance in most of these languages is insufficiently evaluated~\cite{hu2025quantifying,shani2026roots}. This gap is especially pronounced for low-resource languages, which collectively represent a substantial portion of the world’s linguistic diversity.

Evaluating language competence demands more than measuring syntax, semantics and morphology or fluency of the generated text. Language is inseparable from the cultural values, social norms and shared historical experiences \citep{hershcovich-etal-2022-challenges}. Consequently, a reliable language model must internalize these nuances to serve reliably as a tool for information retrieval and content generation. Without such grounding, generated content may appear superficially fluent while remaining semantically unstable, culturally shallow, or socially misaligned.  
The few existing evaluations of LLM creative writing have focused almost exclusively on English. 
\citet{gomez-rodriguez-williams-2023-confederacy}
show that top commercial LLMs match or exceed human
writers on fluency, coherence, and style in English short
fiction, while humans retain an edge in originality and
humour. Whether these conclusions transfer to a
low-resource, culturally distant language is, to our
knowledge, untested.

Beyond English-centric creative writing, prior work on multilingual evaluation has primarily focused on benchmarking general language understanding \cite{ahuja2023mega, xuan2025mmlu} and translation quality \cite{kreutzer2025d, zhu2024multilingual}, mostly using English benchmarks as foundation \cite{thellmann2024towards}. Researchers have reported substantial performance gaps in low-resource languages; however, these studies have largely not examined long-form generative capabilities in depth \cite{unjum2025llms, anikina2025rigorous}. Similarly, research on cultural and value alignment has shown that language models tend to reflect dominant cultural priors, but these studies have largely been limited to classification or preference tasks rather than open-ended narrative generation \cite{tao2024cultural, santurkar2023whose}. While recent work has explored automated evaluation and LLM-as-judge frameworks~\cite{gu2024survey}, these methods have been shown to exhibit reliability issues and sensitivity to evaluation settings, particularly in more complex generation scenarios \cite{schroeder2024can, thakur2025judging}.

In this work, we question how well contemporary multilingual LLMs are grounded in the linguistic and cultural dimensions of low-resource languages. We take Urdu as our case study and investigate the capacity of LLMs to produce stories that are linguistically correct, semantically meaningful, and culturally situated within the Pakistani context. Story generation is a deliberate design choice, unlike factual or short-form tasks, narrative generation demands contextual awareness, structural coherence, and cultural sensitivity sustained over longer text, making it a particularly revealing testbed for model capability.

Concretely, we examine whether LLM-generated Urdu stories preserve contextual relevance, avoid formulaic narrative patterns, and handle culturally sensitive themes appropriately, whether those themes originate from Urdu literary traditions or are adapted from Western sources. This leads us to the following research questions:

\begin{itemize}
\item \textbf{Originality.} Do LLMs generate original Urdu narratives, or reproduce remembered source texts, and how does this reproduction degrade as source-text visibility in training data decreases?
\item \textbf{Linguistic and semantic accuracy.} How accurately do LLMs handle Urdu-specific grammar and do their stories remain semantically coherent at the sentence and discourse level?
\item \textbf{Cultural fidelity.} To what extent do generated stories preserve Pakistani cultural context?
\item \textbf{Narrative structure.} What structural patterns  characterise LLM-generated Urdu stories, and how do they differ across models?
\item \textbf{Tractability.} Which of the above failures can be remediated by targeted prompting?
\end{itemize}

To study these questions, we introduce \textbf{Urdu-Stories}, a dataset of 93 stories generated using three contemporary LLMs: GPT-5.1, Qwen-3-Max, and DeepSeek-3.1. All stories are   {annotated by a native Urdu speaker using a thirteen-label } taxonomy covering linguistic, semantic, {and entity dimensions, with inter-annotator agreement verified by four additional native-speaker annotators on a stratified 60-story subset ($\kappa=0.81$)}. Our analysis reveals five consistent findings. First, models exhibit \textit{memorisation over creativity}: when prompted with well-known literary titles, they frequently reproduce the original story rather than generating a novel narrative. Second, we observe \textit{unnatural title anchoring}: models repeatedly insert the provided title into the story body to signal topical relevance. To quantify this behaviour, we introduce the \textit{Title Repetition Rate (TRR)}, which substantially exceeds the repetition patterns found in human-written stories. Third, we identify \textit{linguistic fragility}, including spelling inconsistencies and grammatical instability. Fourth, models demonstrate \textit{semantic and register instability}, producing factual contradictions, grammatically valid but meaningless sentences, and unprompted switching into Punjabi. Finally, we observe \textit{cultural shallowness}, reflected in pervasive cross-script contamination and a 3.7:1 male-default protagonist bias.

Together, these findings demonstrate that current multilingual LLMs remain fluent in Urdu at the surface level, yet fail to achieve deeper creativity, semantic robustness, and cultural grounding  and, we expect, generalise to many other  low-resource languages , underscoring the need for dedicated development effort beyond surface-level multilinguality.\footnote{We release the dataset, annotations, and evaluation scripts to support future research on open-ended generation in low-resource languages.}

\section{Story Corpus Development}

\subsection{Story Title Selection}
\label{subsec:title_corpus}

We constructed a corpus of 31 story titles drawn
from four different sources
(Table~\ref{tab:title_distribution}): \emph{Translated Short
Story Titles} (TSST) from canonical English short
fiction;
\emph{Urdu Short Story Titles} (UST) from canonical
twentieth-century Urdu writers used \emph{verbatim}
;
\emph{Fables}, short Indo-Pakistani moral tales used
as a memorisation control; and \emph{Neutral Urdu Titles}
(NUT), author-original Urdu titles composed by the
research team as an unseen-prompt control. The complete
title list with authors, sources, and ALA-LC transliterations
is given in Appendix~\ref{app:titles}
(Table~\ref{tab:title_full_list}).

For TSST, we obtained an initial Urdu rendering of each
English title via the GPT-5.1 translation API and then
manually verified every translation with a native Urdu
speaker; published Urdu renderings 
were retained verbatim where they exist.

The four-way split is intentional: TSST tests cross-cultural
transfer, UST tests reproduction vs.\ reinterpretation of
canonical Urdu literature, Fable controls for plot
memorisation, and NUT controls for memorisation in absence
of any prior text. 

Following the leakage-control principle of \citet{gomez-rodriguez-williams-2023-confederacy}, who tested LLMs on an epic combat between Ignatius J.\ Reilly and a pterodactyl, a scenario invented specifically so that no matching text could exist in any model's training data, all six NUT titles were composed by the authors of this paper and do not correspond to any published Urdu literary work or publicly available source. Titles were designed to be contextually plausible within Pakistani settings (e.g.\ \foreignlanguage{arabic}{سوات کی
شام}{, ``Evening in Swat'') without being derivable from any known text, and no information about them was shared publicly before story generation. This ensures that stories generated under NUT prompts reflect the model's generative behavior rather than reproduction of a memorized source.
}

 \subsection{Generation Protocol}
\label{subsec:generation}

Three LLMs were evaluated: \textbf{GPT}
(\texttt{5.1}), \textbf{QWEN} (\texttt{3-Max}),
and \textbf{DeepSeek} (\texttt{3.1}), all accessed
in \texttt{November, 2025}. Each title was issued to every model
exactly once using a fixed zero-shot prompt:

\begin{quote}\itshape
``Write a 2000-word Urdu story in a Pakistani cultural
context. The title is \foreignlanguage{arabic}{دوسری عورت}.''
\end{quote}

\subsection{Corpus Statistics}
\label{subsec:corpus_stats}

The protocol produced 93 stories, totalling 151{,}965 words
and 9{,}460 sentences
(Table ~\ref{tab:corpus_statsistics}). 

\begin{table}[ht]

\centering\small
\caption{Per-model corpus statistics}
\scalebox{0.9}{

\begin{tabular}{lcccc}
\toprule
\textbf{Statistic} & \textbf{GPT} & \textbf{QW} & \textbf{DS} & \textbf{Overall} \\
\midrule
Stories                  & 31      & 31      & 31      & 93      \\
Avg words / story       & 1{,}587 & 1{,}382 & 1{,}934 & 1{,}634 \\
Min         & 726 & 946  & 1{,}461  & 1{,}044  \\
Max words & 1{,}997 &1{,}856&  3{,}030 &  2{,}294\\
Std.\ dev.\ (words)      & 290     & 199     & 317     & 351     \\

Total words              & 49{,}199 & 42{,}827 & 59{,}939 & 151{,}965 \\
Avg. sentences / story   & 122.9    & 101.6    & 149.6   & 101.7   \\
Total sentences          & 3{,}809 & 3{,}149 & 4{,}639 & 11{,}597 \\
Length compliance (\%)   & 58.1    & 12.9    & 80.6    & 50.5    \\
\bottomrule
\end{tabular}}
\label{tab:corpus_statsistics}
\end{table}
\subsection{Annotation and Inter-Annotator Agreement}
\label{subsec:annotation}

A native Urdu speaker annotated all 93 stories in
Label Studio using thirteen labels across three categories
(Table~\ref{tab:labels_compact}). Most labels are applied
at the \emph{sentence} level; entity labels (Character,
City, Protagonist,
) are applied once per
\emph{story}. \emph{Agreement Case} (gender,
oblique-case agreement) is treated as a sub-type of
\emph{Grammar Error}; \emph{Semantic Anomaly} marks sentences that
are syntactically well-formed but semantically empty;
\emph{Context Mismatch} marks factual contradictions.

A stratified subset of 60 stories (20 per model,
balanced across title categories) was independently
re-annotated by four additional native Urdu speakers (15
stories per annotator) . Cohen's $\kappa$ was
computed using exact text+label match for PER/City,
substring-tolerant match for protagonist labels , and same-label-in-same-sentence
match for all other labels. Overall agreement was
$\kappa$~=~\textbf{0.81} (substantial reliability); per-model
and per-label scores are reported in
Appendix~\ref{app:iaa}.

The thirteen-label taxonomy itself was developed through an
iterative inductive process: the annotation team first read a
sample of generated stories, cataloged recurring error
patterns, and progressively refined and merged categories and
labels based on what was actually observed in the data, rather
than adopting a taxonomy from prior theoretical work.

 \begin{table*}[!htbp]
\centering\scriptsize
 \renewcommand{\arraystretch}{0.82}
\setlength{\tabcolsep}{4pt}
 \caption{Annotation labels. G = granularity
(\textbf{U}~story / \textbf{S}~Sentence).}
\label{tab:labels_compact}
\begin{tabular}{p{2.7cm} c p{9.6cm}}
\toprule
\textbf{Label} & \textbf{G} & \textbf{Definition / Example} \\
\midrule
\multicolumn{3}{l}{\textit{\textbf{Linguistic}}} \\
Grammar Error   & S & Syntax \foreignlanguage{arabic}{  احمد اور ثنا نے سب کے لیے تحفے لائے }\\
Agreement Error & S &Obliqueness .
   \foreignlanguage{arabic}{چھوٹی سی لنگر خانہ} $\rightarrow$
   \foreignlanguage{arabic}{چھوٹا سا لنگر خانہ}  \\
Spelling Error  & S & Misspelled Urdu word.
   \foreignlanguage{arabic}{مسئیبت} $\rightarrow$
   \foreignlanguage{arabic}{مصیبت} \\
Space Deletion  & S & Two words fused, no space.
   \foreignlanguage{arabic}{چاہتیہوں} $\rightarrow$
   \foreignlanguage{arabic}{چاہتی ہوں} \\
Em-dash & S & \\
\midrule
\multicolumn{3}{l}{\textit{\textbf{Semantic}}} \\
Context Mismatch & S & Factual contradiction within story.
   ``had no children \ldots her son went to school'' \\
Semantic Anomaly      &    S & Grammatical but meaningless.
   \foreignlanguage{arabic}{آواز میں عجیب کانپ تھی} \\
Cultural Incons.\ & S & Non-Pakistani reference.
   \foreignlanguage{arabic}{چند اسٹیشن کے بعد، ٹرین کا ریفریشمنٹ وین کھلا} \\
Script Contamination     & S & Non-Urdu script token (whole/partial).
   \textit{hospital}; \foreignlanguage{arabic}{ح}aj\foreignlanguage{arabic}{ی} \\
\midrule
\multicolumn{3}{l}{\textit{\textbf{Entities}}} \\
Character              & U & Character name. \foreignlanguage{arabic}{حمزہ} \\
City             & U & Place name. \foreignlanguage{arabic}{لاہور} \\
MaleProtag.      & U & Main male character.
   \foreignlanguage{arabic}{حمزہ} \\
FemaleProtag.    & U & Main female character.
   \foreignlanguage{arabic}{ثنا} \\
\bottomrule
\end{tabular}
\end{table*}

\section{Qualitative Analysis}
Several models generated grammatically correct Urdu but failed to preserve culturally authentic Pakistani dialogues and naming conventions.

\subsection{Linguistic and Semantic Accuracy}

Manual annotation of all 93 generated stories 
identified 
quality issues across three dimensions: 
linguistic errors , semantic 
coherence failures, and Western-influenced cultural 
mismatches . 
Table~\ref{tab:error_summary} presents 
the complete breakdown.

\subsubsection{Linguistic Errors}

Linguistic errors constituted the largest share 
of identified issues. Four subtypes 
were annotated: grammatical mistakes, spelling 
errors, space deletion, and em-dash.

\textbf{Grammatical errors} were the most 
frequent category overall. QWEN exhibited the highest rate 
, followed by 
DeepSeek  and GPT  . Recurring patterns included 
incorrect postposition usage (\foreignlanguage{arabic}{نے}، \foreignlanguage{arabic}{کو}،  
\foreignlanguage{arabic}{سے}), broken verb-subject agreement, and 
malformed compound sentences.
To characterise the kinds of grammatical failures
observed, we discuss three representative examples,
each highlighting a different fault line in Urdu
syntax.

\textit{(i) Ergative case marker omission.}
In QWEN's
\foreignlanguage{arabic}{زرین خاموشی سے بندوق نیچے رکھ دی}
(\textit{Zarīn khāmōshī sē bandūq nīchē rakh dī},
``Zarin quietly put the gun down''), the agent
\foreignlanguage{arabic}{زرین} requires the ergative
postposition \foreignlanguage{arabic}{نے}
(\textit{nē}) because the verb
\foreignlanguage{arabic}{رکھ دی} (\textit{rakh dī}) is
transitive and perfective.
The same omission recurs in GPT's output .
Ergative \foreignlanguage{arabic}{نے} marking is a
hallmark of Urdu morphosyntax with no English analogue,
and its absence suggests cross-lingual transfer from a
nominative-only template.

\textit{(ii) Gender agreement violation.}
In GPT's
\foreignlanguage{arabic}{اس کے منہ سے آج کوئی غرور بھرا بھونک نہیں نکل رہا تھا}
(\textit{Us kē munh sē āj kōʾī ghurūr bharā bhaunk
nahīṅ nikal rahā thā}, ``no proud bark came out of his
mouth today''), the noun
\foreignlanguage{arabic}{بھونک} (\textit{bhaunk},
\emph{bark}) is feminine in Urdu and triggers feminine
agreement on both the modifier and the verb. The model nevertheless propagates masculine
agreement through the entire clause, indicating that
grammatical gender is being assigned by surface
frequency rather than by lexical entry.

\textit{(iii) Aspect mismatch in sequential narrative.}
QWEN's
\foreignlanguage{arabic}{اگلے دن، وہ سویرے اٹھا، نہاتا دھوتا، اوور کوٹ پہنا اور دفتر پہنچ گیا}
(\textit{Aglē din, wuh sawērē uṭhā, nahātā dhōtā, ōvar
kōṭ pahnā aur daftar pahuṅch gayā}, ``the next day, he
got up early, bathed, put on the overcoat, and reached
the office'') chains four events in a single sentence
using past-perfective forms, but inserts
\foreignlanguage{arabic}{نہاتا دھوتا} (\textit{nahātā
dhōtā}, habitual imperfective). This aspectual slip is particularly
diagnostic because it occurs within an otherwise
well-formed sequence, suggesting that the model can
produce isolated verb forms but struggles to maintain
aspectual consistency across a narrative clause.

Their recurrence
across all three models suggests that these are not
random mistakes but reflect a deeper gap in how Urdu
morphosyntax is represented during generation.

\textbf{Agreement Error} A substantial share of grammatical
issues (\textit{cf.}\ Table~\ref{tab:error_summary}) involved
local agreement failures within a 2--3 word window: gender
 agreement(adjective/verb mismatched to the head noun's gender), the most common sub-type; cf.\ example (ii) above) and
\textit{{oblique-case agreement}} {(} a masculine singular noun or
adjective  {failing } to take its oblique form before a postposition e.g.\ \foreignlanguage{arabic}{سر کٹا سوار نے} $\rightarrow$
\foreignlanguage{arabic}{سر کٹے سوار نے}). Both {are mechanical
and } strictly local,  {suggesting } a lightweight post-editing {pass
} over noun-phrase windows could  correct a large share of them
without {retraining the model}.

\textbf{Spelling errors} were distributed across all
three models, suggesting that spelling degradation in
Nastaliq is model-agnostic and likely reflects
tokenisation artefacts rather than model-specific
weaknesses. Four recurring patterns were identified:
hamza misplacement, vowel and diacritic confusion,
visually-similar character substitution, and
English-mediated retroflex confusion. QWEN dropped the
hamza-on-yē in adjectival derivations
(\foreignlanguage{arabic}{دریاوی} for
\foreignlanguage{arabic}{دریائی}, ``riverine''), while
DeepSeek \emph{inserted} a hamza into an Arabic loanword
that does not take one
(\foreignlanguage{arabic}{مسئبت} for
\foreignlanguage{arabic}{مصیبت}, ``calamity''). Vowel-marking errors such as DeepSeek's
\foreignlanguage{arabic}{میچس} for
\foreignlanguage{arabic}{ماچس} (``matchbox'') and GPT's
\foreignlanguage{arabic}{جھلے ہوئے ہاتھ} for
\foreignlanguage{arabic}{جلے ہوئے ہاتھ} (``burnt
hands'')
resemble the mistakes of a learner who has heard a word
but not internalised its written form.

The most diagnostic pattern is the retroflex series
(\foreignlanguage{arabic}{ٹ},
\foreignlanguage{arabic}{ڈ},
\foreignlanguage{arabic}{ڑ}), one of the few phonemic
features of Urdu absent from both Arabic and Persian.
DeepSeek produced \foreignlanguage{arabic}{ربر}
(\textit{rabar}) for \foreignlanguage{arabic}{ربڑ}
(\textit{rabaṛ}, ``rubber''), \emph{erasing} a retroflex
flap that English ``rubber'' lacks; conversely it
rendered the Punjabi-Urdu name
\foreignlanguage{arabic}{اللہ دتّہ} as
\foreignlanguage{arabic}{اللہ ڈتہ}, \emph{inserting} a
retroflex where dental \foreignlanguage{arabic}{د} is
correct, overgeneralising the English romanisation
convention that uses retroflex symbols for South Asian
dentals. The fact that these errors move in opposite
directions, both governed by the model's English
representation of the same Urdu phoneme, suggests that
retroflexes are resolved through an English-mediated
phonetic intermediary rather than from a native Urdu
inventory.

\textbf{Space deletion.} Each model fuses a distinct, tokenizer-specific set of Urdu words into single orthographic units, an error invisible to readers but consequential for whitespace-based NLP pipelines (See Appendix~\ref{app:space_deletion} for details).

\textbf{Em-dash} \label{subsec:auto_errors} Urdu prose 
conventionally does not employ em-dashes (---) 
as structural markers; their 
presence indicates stylistic transfer from 
English writing conventions. Automatic 
detection across all 93 stories identified 
a total of \texttt{623} dash occurrences 
(avg~=~\texttt{6.6} per story), with 
\texttt{21} showing the highest frequency.
Per-model
distribution and per-category breakdown are reported
in Table~\ref{tab:em_dash} (Appendix).
{ {QWEN} } produced the most em-dashes both in
total and per story, consistent with its higher
overall rate of English-stylistic transfer.
\subsubsection{Semantic Coherence}

Semantic coherence failures constituted the second 
largest error category. 

\textbf{Semantic Anomaly.} Sentences that are
syntactically well-formed but semantically impossible or
meaningless.  GPT's \foreignlanguage{arabic}{آگ کے چراغ سے اُبلتی
چائے پی رہا تھا} (``he was drinking tea boiling from a lamp of
fire'') is grammatically perfect but  violates real-world
plausibility.
A subtler pattern is near-synonym confusion:
DeepSeek substitutes  \foreignlanguage{arabic}{بڑھاوا}
\emph{{incitement}} for the intended
\foreignlanguage{arabic}{ترقی} (promotion), yielding 
a syntactically intact but semantically incoherent sentence, a
failure fluent Urdu speakers would not produce (further examples
in Appendix~\ref{app:extended_examples}).

\textbf{Context mismatch.}ِ\label{sec:context_mismatch} Context-mismatch errors arise when a generated
sentence is locally well-formed but conflicts with
earlier narrative content, a real-world fact, or a
culturally established script. The
annotated instances fall into five recurring failure
modes, each of which reveals a different limitation
of current Urdu LLMs.

\textit{Kinship-term misuse} is the dominant
sub-category. Urdu encodes a much finer-grained
kinship lexicon than English, distinguishing
paternal from maternal relations
(\foreignlanguage{arabic}{دادی} \textit{dādī}
``father's mother'' vs \foreignlanguage{arabic}{نانی}
\textit{nānī} ``mother's mother''), nephews by
sibling gender (\foreignlanguage{arabic}{بھتیجا}
\textit{bhatījā} ``brother's son'' vs
\foreignlanguage{arabic}{بھانجا} \textit{bhānjā}
``sister's son''), and address forms by social
relation. Models routinely collapse these
distinctions: in one QWEN story, a brother writes
to his sister calling her son
\foreignlanguage{arabic}{بھانجا} (\textit{bhānjā},
``sister's son'') where
\foreignlanguage{arabic}{بھتیجا} (\textit{bhatījā},
``brother's son'') is required; a DeepSeek story
introduces a maternal-side visit but uses
\foreignlanguage{arabic}{دادی اماں} (\textit{dādī
ammā}, ``paternal grandmother'') for the maternal
grandmother encountered there. Address forms are
similarly over-generalised: QWEN uses
\foreignlanguage{arabic}{ابا جان} (\textit{abbā
jān}, ``father, respectful'') for an elderly
stranger across multiple stories, and a DeepSeek
wife addresses her husband as
\foreignlanguage{arabic}{بیٹا} (\textit{beṭā},
``son''). These errors are consistent with an
English-mediated representation in which
``uncle'', ``nephew'', ``grandmother'', and
``dear'' map onto whichever Urdu term is
statistically most frequent in the training data,
regardless of the discourse context.

\textit{Long-range gender breakdown} forms a second
cluster, mostly observed in DeepSeek  {: once } a female protagonist is
 established, the model  {later } produces verbs and adjectives in
the masculine {(e.g} \foreignlanguage{arabic}{جانتا} {instead of
feminine } \foreignlanguage{arabic}{جانتی}({jāntī}{,
fem.); {), or describes her with
masculine-coded items (a girl } wearing
\foreignlanguage{arabic}{عمامہ}
, a male turban. These are not
local agreement slips but long-range failures in which
 discourse-established gender is lost over several sentences
(further examples in Appendix~\ref{app:extended_examples}).

\textit{Cultural and religious anachronism} also
appeared{: a } story has the remains of a Muslim character
cremated  and given a funeral with an empty coffin, {neither of
which conforms } to Islamic burial practice {, and a QWEN story
conflates the Eid crescent (}\foreignlanguage{arabic}{ہلال}) with
a full moon (\foreignlanguage{arabic}{بدر} . The models have
memorized Urdu vocabulary but not the cultural scripts that
constrain its use (further examples in
Appendix~\ref{app:extended_examples}) .

\textit{Intra-story contradictions} include
references to a daughter ``lost five years ago''
who is nonetheless still being educated in the same
paragraph (QWEN),{and } a mother said to have sewn a
blanket after her son had already left (GPT) . These reveal a
lack of persistent state across sentences in
medium-length narratives.

\textit{Physical and commonsense impossibilities}
round out the category:  a chandelier
inside a bus, books packed inside a thermos, and a wrist-watch from 1947 displaying the date 14
November 1971. The surface fluency of these
sentences masks the absence of a world model.

Although Context Mismatch contributes the small
percentage of any linguistic-error category, its
qualitative profile is the most diagnostic of how
multilingual LLMs fail. The kinship and cultural sub-types
in particular indicate that current models reproduce
Urdu surface forms while operating with an
English-anchored semantic representation.

\subsection{Cultural and Script Fidelity}
 \label{sec:cultural}A small but qualitatively striking sub-set of
context-mismatch errors involves the substitution
of Pakistani objects, institutions, and practices
with their nearest Western analogues, while the
surrounding Urdu prose remains fluent. Although
these instances are too few to dominate any single
model's overall error rate, they form a coherent
pattern that reinforces the cross-lingual
contamination hypothesis raised in our spelling
and punctuation sections. Three sub-types recur.

\textit{Western material culture in Pakistani
settings.} A GPT story set in a Punjabi village
describes the bedroom wall as covered with
\foreignlanguage{arabic}{پیلا وال پیپر}
(\textit{pīlā wāl pepar}, ``yellow wallpaper''), an
interior-decor convention essentially absent from
rural Pakistani homes, which use lime-wash or
distemper paint. Another GPT story has a rickshaw
driver refer to a
\foreignlanguage{arabic}{گیس اسٹیشن}
(\textit{gais isṭeshan}, ``gas station''), an
American term, where Pakistani usage is uniformly
\foreignlanguage{arabic}{پٹرول پمپ}
(\textit{paṭrol pamp}, ``petrol pump'').

\textit{Western institutional and role labels.}
DeepSeek refers to a twelve-year-old grandson as a
\foreignlanguage{arabic}{بارہ سالہ جونیئر}
(\textit{bārah sālah jūniyar}, ``twelve-year-old
junior''), importing the American junior--senior
school-grade classification into a Pakistani village
imam's household, where such labels are not used. The
same model also  introduces a
\foreignlanguage{arabic}{والدین میٹنگ}
(\textit{wāldain mīṭing}, ``parents meeting'') as
a literal translation of the English
``parent--teacher meeting'', whereas the
established Urdu term is
\foreignlanguage{arabic}{پی ٹی ایم}
(\textit{pī ṭī em}).

\textbf{Language Register Violations.} A qualitatively
distinct failure model, observed exclusively in DeepSeek, is  the
substitution of Urdu with Punjabi at the phrase and sentence
level{: }characters switch to Punjabi dialogue within an otherwise
Urdu narrative, a discourse-level failure and a direct violation
of the prompt's explicit Urdu instruction . The pattern suggests
 DeepSeek conflates Pakistani linguistic identity with Punjabi,
the majority regional language, rather than maintaining the
prompted language boundary.

\textbf{Script Contamination in LLM-Generated Urdu Text}
\label{subsec:script_contamination}. Beyond lexical foreign-word intrusion, automated
Unicode-level analysis revealed a more fundamental failure
mode: the embedding of non-Urdu script characters (Latin,
Devanagari, CJK and Cyrillic) within generated Urdu text. A
word was flagged as contaminated when any character in it
fell outside the Urdu Unicode range
(U+0600 to U+06FF) and its standard presentation forms,
excluding digits and punctuation. DeepSeek exhibited the
most severe contamination , followed by QWEN 
and GPT. Contamination took two
structural forms: \textit{pure intrusion}, in which an
entire word belongs to a foreign script, and
\textit{partial contamination}, in which a single word
mixes Urdu with foreign characters and so corrupts an
otherwise valid Urdu token. Latin was the most frequent
contaminant across all three models,
followed by Devanagari, with CJK and Cyrillic appearing
only in DeepSeek. Per-script and
per-model breakdowns, full token lists, and Unicode-level
analysis are in Appendix~\ref{app:script_contamination}.

\subsection{Gender Representation in 
              Protagonist Roles}
\label{subsec:gender_bias}
Across all 93 stories, male protagonists
outnumbered female protagonists 70 to 19 with
an overall 3.7:1 ratio
(Table~\ref{tab:gender}). No model approached gender
parity. QWEN was the most male-skewed (5.2:1), while
GPT and DeepSeek shared a 3.1:1 ratio. 

This disparity is particularly noteworthy given that
generation prompts were gender-neutral and contained
no explicit instruction regarding protagonist gender.
The consistent male-default pattern across all three
models suggests that gender bias is encoded in the
underlying generative behaviour rather than being
prompt-induced. This
finding aligns with prior work on gender bias in
neural text generation
\citep{genderbiasllmgenerated2025,VANBLERCK2025100972}.

\subsection{Originality: Source Engagement and Memorisation}
\subsubsection{Title Repetition and Three Modes of Source Engagement}
\label{subsec:title_repetition}

A natural question raised by the linguistic and cultural
errors of the previous sections is whether the underlying
problem is one of \emph{language coverage} or \emph{text
coverage}: does the model fail because it cannot speak Urdu
well, or because it has never read the specific Urdu texts
that our titles reference? To disentangle these, we
introduce a simple, length-invariant lexical diagnostic,
the \textbf{Title Repetition Rate (TRR)}, defined as the
number of exact title-phrase occurrences in the body of a
generated story (excluding the opening line) per 100~words
of body text. Both title and body are normalised before
matching by removing diacritics, applying character-variant
normalisation, stripping punctuation and normalising
whitespace. The metric requires no LLM judge and no
content-level annotation, making it robust to the
methodological concerns that affect more semantic
evaluation pipelines.

Table~\ref{tab:trr_category} shows mean TRR by category
and model. The ordering is identical for every model and
spans a tenfold range, revealing three distinct modes of
source engagement:

\textbf{Plot Reproduction (PR), Fables.}
    For globally circulated fables present in every
    model's training data, TRR is near zero [$\approx$~0.06]. The
    model retrieves a complete memorised plot and does not
    need to re-anchor on the title.

\textbf{Thematic Reproduction (TR), English Classic }
    For Western literary classics translated into Urdu, TRR is
    intermediate (TRR~$\approx$~0.45). The model invokes a remembered
    \emph{theme} (mental-health breakdown, a wife's brief
    liberation, sudden inversion of fortune) without
    retrieving a full plot, and uses the title as a
    partial anchor.

\textbf{Lexical Title Unpacking (LT), Urdu canon}
    For titles drawn from the Urdu literary canon e.g. 
    (\foreignlanguage{arabic}{لحاف},
    \foreignlanguage{arabic}{اوور کوٹ}), texts largely
    absent from web pretraining corpora, TRR rises
    sharply(~$\approx$~0.58, peak~2.79) . With neither plot nor theme to retrieve, the
    only object the model can anchor on is the surface
    meaning of the title itself.

 We call this the Title Lexicalisation Effect: in
the absence of source-text knowledge, an LLM treats a
metaphorical title as a literal writing prompt. The most
striking instance is Chughtai's
\foreignlanguage{arabic}{لحاف}, a 1942 short story
metaphorising hidden lesbian desire, rendered by QWEN as a
literal narrative about a quilt with the word
\foreignlanguage{arabic}{لحاف} (blanket) repeated 35 times
(TRR~=~2.79). Bedi's \foreignlanguage{arabic}{گرم کوٹ}(warm coat)
becomes a story about how to obtain a warm garment, and
Manto's \foreignlanguage{arabic}{ٹھنڈا گوشت}(cold meat), a story of
partition-era trauma, becomes a story about cold meat.

Because TRR is computed on raw text rather than plot
content, the same metric applies to the original Urdu
stories. Source texts of
\foreignlanguage{arabic}{لحاف}(blanket),
\foreignlanguage{arabic}{اوور کوٹ}(over coat),
\foreignlanguage{arabic}{گرم کوٹ} (warm coat)and
\foreignlanguage{arabic}{ٹھنڈا گوشت} (cold meat) 
yield a human-author TRR baseline of 0.04 to 0.52: the
original authors use their title words sparingly, as
thematic anchors rather than literal subjects. LLM TRR
exceeds the human baseline in 8 of 9 comparisons (88.9\%),
with ratios from 1.4$\times$ to 12.7$\times$. The two
extremes (QWEN's \foreignlanguage{arabic}{لحاف} at
5.3$\times$ Chughtai's original and DeepSeek's
\foreignlanguage{arabic}{ٹھنڈا گوشت} at 12.7$\times$
Manto's) are precisely the canonical Urdu texts most
likely to be absent from training data. Per-title
breakdowns and top repetition cases are in
Appendix~\ref{app:TPR}.

\begin{table}[ht]
\caption{Mean TRR by category and model. The tenfold spread
between Fable (0.06) and Urdu Literature (0.58) is the
quantitative signature of the three modes.}
\label{tab:trr_category}
\centering\small
\begin{tabular}{lcccc}
\hline
\textbf{Category} & \textbf{DS} & \textbf{GPT} &
\textbf{QWEN} & \textbf{Avg} \\
\hline
Fable \hfill (PR)              & 0.11 & 0.01 & 0.04 & 0.06 \\
Neutral Urdu \hfill (OR)       & 0.29 & 0.36 & 0.24 & 0.30 \\
English Literature \hfill (TR) & 0.51 & 0.46 & 0.39 & 0.45 \\
Urdu Literature \hfill (LT)    & 0.57 & 0.51 & 0.67 & \textbf{0.58} \\
\hline
\textbf{Overall}               & 0.43 & 0.40 & 0.40 & 0.41 \\
\hline
\end{tabular}
\end{table}

\subsubsection{Memorisation on Known Source Texts}
\label{subsec:memorization}

The Lexical Title Unpacking mode of
Section~\ref{subsec:title_repetition} has no source plot to
memorise. This subsection asks the complementary question:
when a source text \emph{is} plausibly in training data
(fables and Western classics), how much of its plot, theme
and resolution does the model reproduce? For every
\emph{(original, generated)} pair in our corpus ($n=12$
titles per model, comprising 4~fables, 3~English classics
and 5~Urdu-canon controls), GPT-5.4 
scores plot, theme, character and resolution overlap on
$[0,1]$. A story is classified as memorisation when at
least two of plot, theme and resolution pass fixed
thresholds; crossed with character overlap, this yields the
2$\times$2 typology of \{Faithful~Retelling,
Disguised~Memorisation, Spin-off, Original\}. Method,
thresholds and the full per-story table are in
Appendix~\ref{app:memorisation}.

Table~\ref{tab:memorisation_results} reports per-model
outcomes. GPT and DeepSeek produce \textit{Faithful
Retellings} in 25\% of cases; QWEN in only 8\%. The composite score (mean of all four overlap axes)
gives the same ordering: GPT~0.50~$>$~DeepSeek~0.43~$\gg$
QWEN~0.27. All 15 Urdu-canon rows receive the
\emph{Original} label across all three models with plot
overlap effectively zero, quantitatively confirming the
TRR-section prediction that the Urdu canon is absent from
training data.

Fables almost always trigger Faithful Retelling, with GPT
at 3/4, DeepSeek at 2/4 and QWEN at 1/4. The one fable all
three models reproduce identically is \emph{The Tortoise
and the Hare} (plot overlap 1.00 for every model). The
single fable nobody reproduces is \emph{The Fool Donkey}
(\foreignlanguage{arabic}{بیوقوف گدھا}), which exists in
multiple narrative variants across South Asian and global
oral tradition. We read its absence as evidence that
coverage in training data is title-specific: when a fable
circulates with a stable canonical plot it is memorised,
but when it circulates as a family of variants no single
version dominates the training signal. English classics
produce a third pattern of high theme overlap, moderate
plot overlap, and low character overlap. This is the
structural signature of a model that has read \emph{about}
the story (summaries, critical commentary) without
memorising the story itself. GPT's \emph{Monkey's Paw}
(theme 0.90, plot 0.29) and DeepSeek's \emph{Story of an
Hour} (plot 0.86, theme 0.95, character 0.75, the only
English-classic PR in our corpus) illustrate the two
extremes of this regime; QWEN on the same titles scores
theme 0.20 and 0.60 respectively, indicating it has neither read nor read-about these works.

Together, the two analyses describe a single gradient: as source-text presence diminishes, LLM Urdu generation degrades gracefully from faithful reproduction ( low TRR, high overlap), through thematic paraphrase (moderate TRR, theme-only overlap), to lexical title-unpacking (high TRR, near-zero overlap). This behaviour is also consistent with recent findings on the “Artificial Hivemind” effect in open-ended LLM generation, where models converge toward retrieval-driven and structurally similar outputs despite the absence of a single correct response \citep{jiang2026artificial}
\begin{table}[ht]
\centering\small
\caption{Per-model memorisation outcomes. \emph{Composite} = mean of plot, theme, resolution
and character overlap.}
\label{tab:memorisation_results}
\begin{tabular}{lccc}
\hline
\textbf{Outcome (\%)} & \textbf{GPT} & \textbf{DS} & \textbf{QWEN} \\
\hline
Faithful Retelling     & 25.0 & 25.0 & \phantom{0}8.3 \\
Original               & 58.3 & 66.7 & 83.3 \\
Spin-off               & 16.7 & \phantom{0}8.3 & \phantom{0}8.3 \\
\hline
Composite score        & \textbf{0.50} & 0.43 & 0.27 \\
\hline
\end{tabular}
\end{table}

\subsection{Narrative Structure: Character Networks}
\label{sec:networks}

During annotation it was observed that the generated
stories felt unusually warm, with QWEN's stories feeling
the warmest. To test whether this impression reflected a
measurable structural signal, the signed character
network methodology of~\citep{networkpaper2025} was
applied to our corpus. For every story, GPT-5.4 extracted
the characters and their pairwise relationships, each
labelled as positive, negative or neutral with intensity
1 to 3. A signed weighted graph was built per story.
Extraction prompt, per-story
distributions, and the rationale for excluding the human
corpus are in Appendix~\ref{app:network_details}.  {For a
story's signed graph $G=(V,E)$ with edge weight
$w_e\in\{-3,\ldots,+3\}$: }\emph{{density}}{~$=2|E|/(|V|(|V|-1))$;
}\emph{{clustering}} {is the mean local clustering coefficient;
}\emph{{positivity ratio}} {is the share of non-neutral edges that
are positive; and }\emph{{mean signed weight}} {rescales
$\frac{1}{3|E|}\sum_e w_e$ to $[-1,+1]$.
} 

\paragraph{Small, tightly bound social worlds.}
Ignoring neutral acquaintances, the positivity ratio is
0.78 for GPT and 0.82 for both DeepSeek and QWEN. The
median is 1.00 for GPT and QWEN and 0.92 for DeepSeek,
meaning that the majority of generated stories contain
\emph{no} negative relationship at all
(Table~\ref{tab:network_metrics}). One of the \textit{GPT} story illustrates the pattern: of eleven characters and ten relationships, nine are positive (a doctor mentors the protagonist, a friend invites him to a gathering, a religious teacher offers spiritual guidance, his father loves him, his mother prays for him, even a passing elderly woman blesses him), and the single negative edge is confined to the explicitly antagonistic role. This protagonist-cushioning structure recurs across all three models. The annotators' QWEN $>$ GPT $>$ DeepSeek ranking is only partially supported: QWEN sits at the top of the warmth scale, but DeepSeek is statistically indistinguishable, and the ranking collapses into a coarser \{QWEN, DeepSeek\} $>$ GPT split.

Intensity separates the models more cleanly than
positivity does. The mean signed edge weight on
$[-1,+1]$ is 0.29 for GPT against 0.37 for DeepSeek and
0.38 for QWEN: QWEN and DeepSeek dial relationships up
to maximum intensity (mentor-as-spiritual-guide,
mother-as-saint), while GPT distributes more weight at
intensities 1 and 2, producing supportive but ordinary
relationships. This is consistent with our earlier
finding (Section~\ref{subsec:memorization}) that GPT and
DeepSeek anchor most strongly to remembered source plots:
memorisation and emotional amplification are correlated
tendencies, and DeepSeek ranks high on both.

\paragraph{Same prompt, different casts.}
Despite identical Urdu titles, the three models populate
their stories with near-disjoint character sets. Pairwise
Jaccard overlap on character names is 0.013 to 0.018
across model pairs, with median exactly zero: for more
than half of all prompts, no two models share a single
character name. The structural patterns above are
therefore not driven by a shared casting template; each
model explores a distinct region of the character space.

\begin{table}[h]
\centering\small
\caption{Network metrics per model (mean over 31~stories;
GPT-5.4 extraction). Signed-only rows restrict to
stories with $\geq 1$ non-neutral edge.}
\label{tab:network_metrics}
\begin{tabular}{lccc}
\hline
\textbf{Metric} & \textbf{GPT} & \textbf{DS} & \textbf{QWEN} \\
\hline
Characters / story          & 6.3  & 6.0  & 5.7  \\
Edges / story               & 6.8  & 7.6  & 6.6  \\
Density                     & 0.53 & 0.56 & 0.55 \\
Clustering                  & 0.40 & 0.54 & 0.46 \\
Positivity (all)            & 0.78 & 0.82 & 0.82 \\
Mean signed weight (all)    & 0.29 & 0.37 & 0.38 \\
Positivity (signed only)    & 0.80 & 0.82 & 0.82 \\
Mean signed weight (signed) & 0.30 & 0.37 & 0.38 \\
\hline
\multicolumn{4}{l}{\textit{Pairwise character Jaccard (mean / median)}} \\
GPT \& DS                   & \multicolumn{3}{c}{0.013 / 0.000} \\
GPT \& QWEN                 & \multicolumn{3}{c}{0.018 / 0.000} \\
DS \& QWEN                  & \multicolumn{3}{c}{0.013 / 0.000} \\
\hline
\end{tabular}
\end{table}

 \subsection{{Robustness and Validation}}
\label{sec:robustness}

{Four additional checks probe the robustness of the findings
above; full tables are in
Appendices~\ref{app:robustness_full}--\ref{app:human_validation}.
}

\paragraph{{Repeated generations.}} {Five titles (one per category)
were regenerated twice more per model (30 additional stories) to
estimate within-condition variance. TRR's SD is modest
(0.027--0.055) and its direction is stable across all nine
model$\times$run comparisons; positivity ratio is the most
stable metric (SD 0.032--0.048, all nine values $>$0.71); the
male-protagonist skew and the linguistic/semantic error patterns
both persist in every run.
}

\paragraph{{English-language control.}} {The same three models
generated the same seven TSST titles in English (21 stories).
Positivity ratio is 17 points higher in Urdu than English (0.807
vs.\ 0.637) and TRR is 3.5--5.8$\times$ higher, while faithful
retelling is }\emph{{lower}} {in Urdu (19.4\% vs.\ 38.1\%) --
consistent with the Urdu memorization gap reflecting thinner
training coverage of the Urdu canon rather than a general model
property.
}

\paragraph{{Cross-family judge validation.}} {Re-scoring
memorization and character networks with Gemini-3.5-Flash as an
out-of-family judge preserves the direction and model ordering
of every headline finding; categorical faithful-retelling
percentages are identical across judges.
}

\paragraph{{Human validation of memorisation labels.}} {A native
Urdu speaker manually re-checked the GPT-4o-mini judge's label on
13 story pairs, agreeing on 10/13 (76.9\%); every disagreement
involved over-estimated plot reproduction on UST titles.
}

 \section{Tractability via LLM Remediation}
\label{sec:tractability}

To test which of the failures documented above can be
corrected without retraining, a targeted post-editing
experiment was conducted on a stratified subset of 15
stories (5 per model, balanced across title categories).
Each story was passed through GPT-5.5 
with a JSON-response prompt instructing it to detect and
minimally fix five error types: grammar errors,
agreement, agreement errors, spelling errors, space
deletion, and semantic anomaly. The prompt enforced
single-token edits where possible (e.g.\ inserting one
missing space, replacing one misspelled word) and
forbade sentence-level rewrites or reformatting. The
full system prompt is given in
Appendix~\ref{app:tractability_prompt}.

GPT-5.5 output was compared against the human annotations
on the same 15 stories. Alignment with human-flagged
errors reached approximately 90\% across the four
linguistic categories: grammar, agreement, spelling and
space deletion were detected and corrected at high
precision, with single-character or single-word fixes
that preserved the surrounding text. Semantic anomaly
(syntactically well-formed but semantically empty
sentences) was detected at lower but useful rates, with
GPT-5.5 successfully repairing cases such as near-synonym
confusion (\foreignlanguage{arabic}{بڑھاوا} for
\foreignlanguage{arabic}{ترقی}) where the lexical
substitution was local. A small number of kinship-term
errors (e.g.\ \foreignlanguage{arabic}{خالہ} for
\foreignlanguage{arabic}{پھوپھی}) were also caught,
indicating partial sensitivity to Urdu's relational
lexicon.

The remediation boundary is sharp. Errors that GPT-5.5
did \emph{not} catch are precisely the failure modes
identified in
Sections~\ref{sec:context_mismatch} and~\ref{sec:cultural}: 
long-range gender breakdown across discourse, cultural
and religious anachronisms (a Muslim character cremated,
a full moon on Eid, Western-context substitutions  {e.g.\ a gas station, refreshment van} ) and Punjabi
register switching in DeepSeek output. These errors
require background knowledge of Pakistani cultural
scripts, persistent discourse state, or commitment to a
prompted language register, none of which a sentence-level
proofreading pass can supply.

\section{Conclusion}

This study evaluates whether contemporary multilingual LLMs are
genuinely competent in Urdu or only fluent on its
surface. A corpus of LLM generated Urdu-Stories  annotated by
native speakers, give a consistent answer. When a
story's source text is well represented online, models
reproduce it; when it is not, they fall back on
unpacking the title itself. 
The surface language is fluent but not native, with
errors clustering on features English does not share:
ergative case, gender agreement, retroflex consonants
spelled through an English intermediary. Father's sister
become Mother's sister, Muslim characters are cremated,
gas stations appear in Punjabi villages. The social
worlds are oddly warm, with the median story containing
no negative relationship at all and male protagonists
outnumbering female ones almost four to one. A modern
model can proofread away most of the grammar and
spelling errors, but it cannot fix the cremation or the
gas station. The models can spell and form sentences,
mostly, but they cannot yet write a Pakistani story.
\section*{Limitations}
\label{subsec:trr_limit}

The Urdu-Stories corpus, comprising 93 stories generated across three commercial models for a single language, remains modest in scale. Extending the analysis to additional commercial and open-weight systems (e.g., Llama, Aya) and to other low-resource languages would strengthen the generality of the cross-model findings.

Our comparison against human-authored Urdu literature relies on published canonical short fiction rather than prompt-matched human generations. We intentionally treat these literary works as culturally validated reference distributions for Urdu narrative structure, thematic development, and stylistic behaviour. Metrics such as Title Repetition Rate (TRR) are length-normalised specifically to permit comparison between LLM-generated and human-authored texts despite differences in story length and authorship conditions. While prompt-controlled human generations could provide a more experimentally matched baseline, published Urdu literature constitutes a stronger benchmark for evaluating whether LLMs reproduce the linguistic, cultural, and stylistic properties of native literary writing.

Exact-phrase TRR may underestimate title anchoring when a model decomposes a multi-word title into its constituent lexical items. GPT’s \foreignlanguage{arabic}{گرم کوٹ} (TRR~=~0.16) illustrates this behaviour: although the full phrase appears only three times, the head noun \foreignlanguage{arabic}{کوٹ} recurs throughout the story, with the protagonist—wrapped in a \foreignlanguage{arabic}{شال}—repeatedly longing for a \foreignlanguage{arabic}{گرم کوٹ}. The title therefore remains thematically central even when the exact surface form is infrequent. Partial-word, lemma-level, and synonym-aware matching are natural extensions that we leave to future work. Moreover, commercial API outputs are non-deterministic and model versions evolve over time, limiting exact reproducibility.

{We attempted to include an open-weight multilingual model
(LLaMA-70B) for comparison, but in this long-form Urdu
story-generation setting it entered repetitive generation loops
after approximately 500 words, producing outputs unsuitable for
analysis; extending the comparison to a broader range of
open-weight multilingual models is left to future work.
}

\bibliography{custom}

\appendix

\section{Appendix}

\begin{table}[ht]

\caption{Protagonist Gender Distribution by Model}
\label{tab:gender}
\scalebox{0.9}{
\begin{tabular}{lcccc}
\hline
\textbf{Model} &
\textbf{Male} &
\textbf{Female} &
\textbf{Total} &
\textbf{M:F Ratio} \\
\hline
DeepSeek   & 22 & 7 & 29 & 3.1:1 \\
GPT        & 22 & 7 & 29 & 3.1:1 \\
QWEN       & 26 & 5 & 31 & 5.2:1 \\
\hline
\textbf{Overall} & \textbf{70} & \textbf{19} & \textbf{89} & \textbf{3.7:1} \\
\hline
\end{tabular} }
\end{table}
\label{sec:appendix}
\begin{table*}[htbp]
\centering
\caption{Manual Annotation Counts by Label and Model}
\scalebox{0.9}{
\begin{tabular}{llcccc}
\hline
\textbf{Category} &
\textbf{Label} &
\textbf{GPT} &
\textbf{QWEN} &
\textbf{DS} &
\textbf{Total} \\
\hline
\multirow{4}{*}{Linguistic}
  & Grammar Error & 72  & 138 & 106  & 316 \\
  & Agreement Error     & 41   & 92   & 39   & 172   \\
  & Spelling Error      & 16  & 40  & 52  & 108  \\
  & Space Deletion      & 79  & 11   & 50  & 140  \\
  & Em-Dash      & 261  & 320   & 42  & 623  \\

\cline{2-6}
  & \textbf{Subtotal}   & \textbf{469} & \textbf{601} 
                        & \textbf{289} & \textbf{1359} \\
\hline
\multirow{4}{*}{Semantic}
 
  & Context Mismatch    & 8  & 21  & 21  & 50  \\
  & Semantic Anomaly             & 41  & 92  & 39   & 172  \\
  & Western Influenced Cultural Mismatch            & 3  & 0  & 5   & 8  \\
  & Script Contamination             & 11  & 14  & 61   & 86  \\
  
\cline{2-6}
  & \textbf{Subtotal}   & \textbf{63} & \textbf{127} 
                        & \textbf{126} & \textbf{316} \\
\hline

\hline
\multicolumn{2}{l}{\textbf{Total Errors}}
  & \textbf{532} & \textbf{728} 
  & \textbf{415} & \textbf{1675} \\
\hline
\multirow{4}{*}{Entities}
  & Character Name                & 94  & 80  & 99  & 273 \\
  & City                & 40  & 49  & 37  & 126  \\
  & MaleProtag     & 22   & 26   & 22   & 70   \\
  & FemaleProtag     & 7   & 5   & 7   & 19   \\
\cline{2-6}

\hline
\end{tabular}
}

\label{tab:error_summary}
\end{table*}

\begin{table*}[htbp]
\centering\small
\caption{Em-dash usage per model.}
\label{tab:em_dash}
\begin{tabular}{lcccc}
\toprule
\textbf{Model} & \textbf{Total} & \textbf{Avg} &\textbf{Min} &\textbf{Max} \\
\midrule
GPT      & 261 & 8.4 & 0 & 18 \\
QWEN     & 320 & 10.32 & 3&21 \\
DeepSeek & 42 & 1.35 & 0&17 \\
\midrule
\textbf{Overall} & \textbf{623} & \textbf{6.7} & 
\textbf{0}&
\textbf{21} \\
\bottomrule
\end{tabular}
\end{table*}

\section{Title Repetition: Method and Extended Results}
\label{app:TPR}

\subsection{Metric Definition}
\label{app:trr_method}

The Title Repetition Rate is defined as

\begin{equation}
    \mathrm{TRR} =
        \frac{\mathrm{count}(\textit{title\_phrase}
                             \in \textit{story\_body})}{\textit{total\_words}} \times 100,
\end{equation}

where the title phrase is matched exactly against the
normalised body of the generated story, excluding the
opening line. Normalisation removes diacritics, applies
standard Urdu character-variant folding (e.g.\ different
forms of \foreignlanguage{arabic}{ی} and
\foreignlanguage{arabic}{ہ}), strips punctuation, and
collapses internal whitespace. Both title and body are
normalised identically before matching.

\subsection{Per-Model Aggregate Statistics}
\label{app:trr_overall}

Table~\ref{tab:trr_overall} reports per-model aggregates
across all 93 stories. DeepSeek produces the longest
stories on average and also the highest absolute count of
title repetitions per story; QWEN and GPT are close on
TRR despite QWEN producing the shortest stories and GPT
the most TRR~=~0 stories.

\begin{table}[h]
\caption{Per-model TRR aggregates across all 93 stories.
\textit{Avg Count} = mean exact title occurrences per
story; \textit{Avg Length} = mean story length in words;
\textit{TRR=0} = number of stories with no title
mention.}
\label{tab:trr_overall}
\centering\small
\scalebox{0.9}{
\begin{tabular}{lcccc}
\hline
\textbf{Model} & \textbf{Avg TRR} & \textbf{Avg Count}
 & \textbf{Avg Length} & \textbf{TRR=0} \\
\hline
GPT      & 0.40 & 6.06 & 1{,}584 & 5 \\
QWEN     & 0.40 & 5.19 & 1{,}379 & 6 \\
DeepSeek & 0.43 & 8.12 & 1{,}908 & 2 \\
\hline
\textbf{Overall} & 0.41 & 6.46 & 1{,}624 & 13 \\
\hline
\end{tabular}
}
\end{table}

\begin{figure}[h]
    \centering
    \includegraphics[width=\linewidth]
        {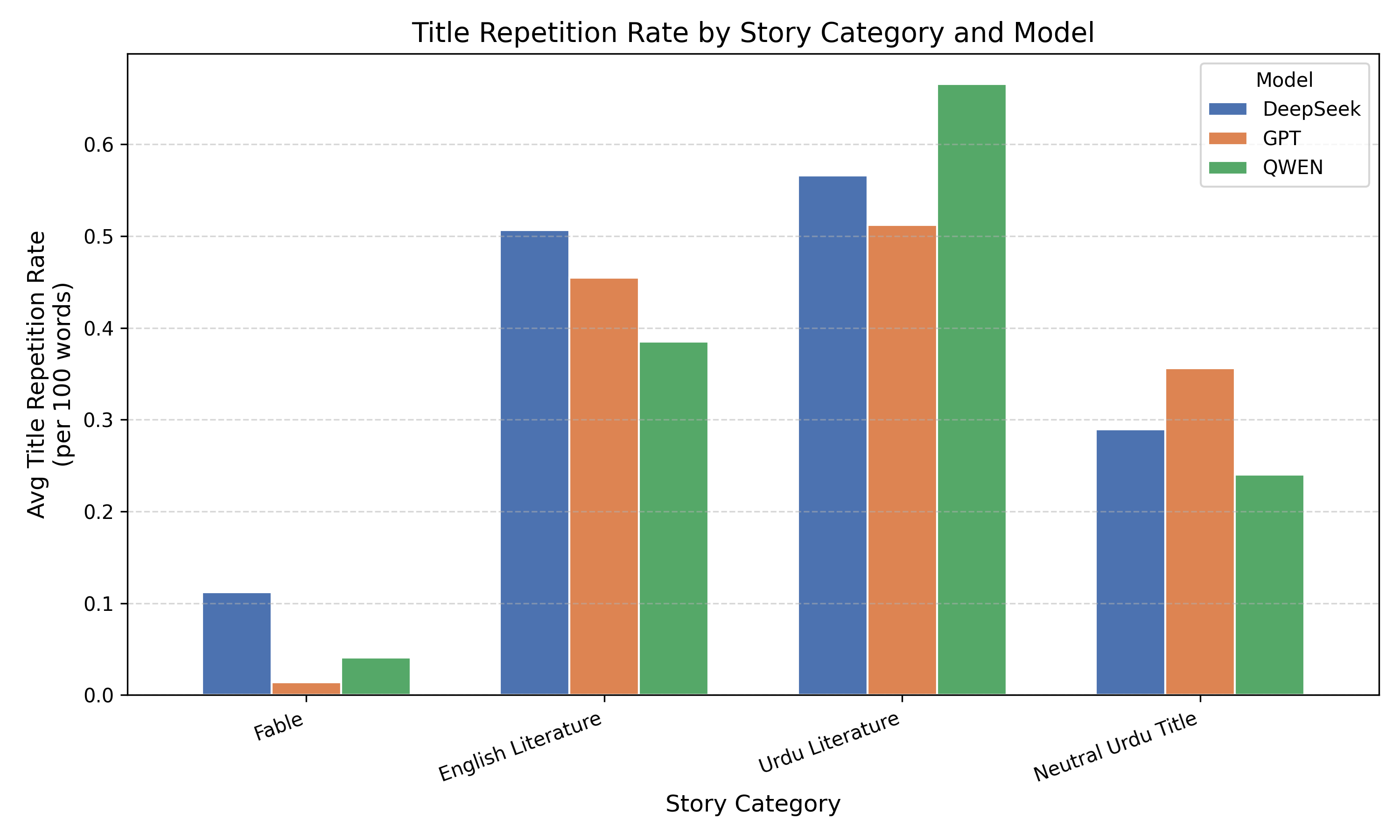}
    \caption{Average TRR by story category and model. 
    Urdu Literature titles show consistently highest TRR, 
    reflecting systematic literal interpretation of 
    metaphorical titles.}
    \label{fig:trr_category}
\end{figure}

\section{Full Story Title Corpus}

\label{app:titles}

\begin{table}[h]
\centering\small
\caption{Distribution of story titles across the four
source categories.}

\label{tab:title_distribution}
\begin{tabular}{lcc}
\toprule
\textbf{Source} & \textbf{Abbrev.} & \textbf{\#} \\
\midrule
Translated Short Story Titles & TSST  & 10  \\
Urdu Short Story Titles       & UST   & 11 \\
Fables                        & Fable & {4  }\ \\
Neutral Urdu Titles           & NUT   & 6  \\
\midrule
\textbf{Total}                &       & \textbf{31} \\
\bottomrule
\end{tabular}

\end{table}
Table~\ref{tab:title_full_list} lists all 31 titles with
their original authors (where applicable), source category,
and ALA-LC transliteration. Per-title generation statistics
appear in Table~\ref{tab:avg_repetition_cases}.

\begin{table*}[htbp]
\centering\scriptsize
\renewcommand{\arraystretch}{1.2}
\caption{Complete list of 31 story titles used in this study.
TSST = Translated Short Story Title (English origin);
UST = Urdu Short Story Title (canonical Urdu/South-Asian
literature); Fable = traditional moral tale;
NUT = Neutral Urdu Title (composed by authors).}
\label{tab:title_full_list}
\begin{tabular}{c p{4.2cm} p{3.5cm} p{3.8cm} c}
\toprule
\textbf{ID} & \textbf{Title (Urdu)} &
\textbf{Transliteration / English} &
\textbf{Original Author / Source} & \textbf{Cat.} \\
\midrule
1  & \foreignlanguage{arabic}{ایمان کا سفر}      & Īmān kā Safar / Journey of Faith       & --- (composed)              & NUT   \\
2  & \foreignlanguage{arabic}{آخری ٹرین}         & Ākhrī Ṭrēn / The Last Train             & --- (composed)              & NUT   \\
3  & \foreignlanguage{arabic}{سوات کی شام}       & Swāt kī Shām / Evening in Swat          & --- (composed)              & NUT   \\
4  & \foreignlanguage{arabic}{آخری بس}           & Ākhrī Bas / The Last Bus                & --- (composed)              & NUT   \\
5  & \foreignlanguage{arabic}{عید کا نیا چاند}   & ʿĪd kā Nayā Chānd / The New Moon of Eid & --- (composed)              & NUT   \\
6  & \foreignlanguage{arabic}{ایمان کی تلاش}     & Īmān kī Talāsh / Search for Faith       & --- (composed)              & NUT   \\
7  & \foreignlanguage{arabic}{پیلا وال پیپر}     & Pīlā Vāl Pēpar / The Yellow Wallpaper   & Charlotte P.\ Gilman        & TSST  \\
8  & \foreignlanguage{arabic}{لاٹری}             & Lāṭrī / The Lottery                     & Shirley Jackson             & TSST  \\
9  & \foreignlanguage{arabic}{بندر کا پنجہ}      & Bandar kā Panjah / The Monkey's Paw     & W.\ W.\ Jacobs              & TSST  \\
10 & \foreignlanguage{arabic}{دوسری عورت}        & Dūsrī ʿAurat / The Second Woman         & (Urdu literary tradition)   & UST   \\
11 & \foreignlanguage{arabic}{ٹھنڈا گوشت}        & Ṭhanḍā Gōsht / Cold Meat                & Saadat Hasan Manto          & UST   \\
12 & \foreignlanguage{arabic}{خاموش اذان}        & Khāmōsh Az̤ān / Silent Call to Prayer   & --(composed)  & UST   \\
13 & \foreignlanguage{arabic}{خرگوش اور کچھوا}   & Khargōsh aur Kachhwā / Hare and Tortoise & Aesop / Folk                & Fable \\
14 & \foreignlanguage{arabic}{پیاسا کوا}         & Piyāsā Kawwā / The Thirsty Crow         & Aesop / Folk                & Fable \\
15 & \foreignlanguage{arabic}{بیوقوف گدھا}       & Bēwaqūf Gadhā / The Foolish Donkey      & Folk                        & Fable \\
16 & \foreignlanguage{arabic}{اندھا آدمی اور ہاتھی} & Andhā Ādmī aur Hāthī / Blind Men \& Elephant & John G.\ Saxe          & TSST  \\
17 & \foreignlanguage{arabic}{لحاف}              & Liḥāf / The Quilt                       & Ismat Chughtai              & UST   \\
18 & \foreignlanguage{arabic}{جلا وطن}           & Jalā Waṭan / Exiled                     & Ghulam Abbas   & UST   \\
19 & \foreignlanguage{arabic}{آگ کا دریا}        & Āg kā Daryā / River of Fire             & Qurratulain Hyder           & UST   \\
20 & \foreignlanguage{arabic}{گرم کوٹ}           & Garm Koṭ / Warm Coat                    & Rajinder Singh Bedi         & UST   \\
21 & \foreignlanguage{arabic}{اوور کوٹ}          & Ōvar Koṭ / The Overcoat                 & Ghulam Abbas                & UST   \\
22 & \foreignlanguage{arabic}{لالچی کتا}         & Lālchī Kuttā / The Greedy Dog           & Aesop / Folk                & Fable \\
23 & \foreignlanguage{arabic}{پچھتاوا}           & Pachtāwā / Regret                       & (Urdu literary tradition)           & UST  \\
24 & \foreignlanguage{arabic}{سر کٹا سوار}       & Sar Kaṭā Sawār / The legend of Sleepy Hollow     & Washington Irving           & TSST  \\
25 & \foreignlanguage{arabic}{ایک گھنٹے کی کہانی} & Ek Ghanṭē kī Kahānī / Story of an Hour & Kate Chopin                 & TSST  \\
26 & \foreignlanguage{arabic}{پہاڑی کی برف}      & Pahāṛī kī Barf / The Snows of Kilimanjaro   & Ernest Hemingway            & TSST  \\
27 & \foreignlanguage{arabic}{لڑکے اور لڑکیاں}   & Laṛkē aur Laṛkiyāṅ / Boys and Girls     & Alice Munro                 & TSST  \\
28 & \foreignlanguage{arabic}{سلطانہ کا خواب}    & Sulṭānah kā Khwāb / Sultana's Dream     & Rokeya S.\ Hossain          & UST   \\
29 & \foreignlanguage{arabic}{خاتون یا شیر}      & Khātūn yā Shēr / The Lady or the Tiger  & Frank R.\ Stockton          & TSST  \\
30 & \foreignlanguage{arabic}{خط}                & Khaṭ / The Letter                       & Anton Chekhov               & TSST  \\
31 & \foreignlanguage{arabic}{پت جھڑ کی آواز}    & Pat Jhaṛ kī Āwāz / Sound of Autumn Leaves & Qurratulain Hyder& UST   \\
\bottomrule
\end{tabular}
\end{table*}

\begin{table*}[htbp]
\centering

\caption{
Average Title Repetition Rates Across Stories.
TSST = Translated Short Story Title,
UST = Urdu Short Story Title,
NUT = Neutral Urdu Title.
}

\label{tab:avg_repetition_cases}

\scriptsize
\renewcommand{\arraystretch}{1.15}

\begin{tabular}{c p{5cm} c c c}
\hline

\textbf{Story ID} &
\textbf{Title (ALA-LC Transliteration / English Translation)} &
\textbf{Source} &
\textbf{Avg.\ TRR} &
\textbf{Avg.\ Count} \\

\hline

1 &
\makecell[l]{
\foreignlanguage{arabic}{ایمان کا سفر} \\
(Īmān kā Safar / Journey of Faith)
}
& NUT & 0.2459 & 3.6667 \\

2 &
\makecell[l]{
\foreignlanguage{arabic}{آخری ٹرین} \\
(Ākhrī Ṭrēn / The Last Train)
}
& NUT & 0.6595 & 8.6667 \\

3 &
\makecell[l]{
\foreignlanguage{arabic}{سوات کی شام} \\
(Swāt kī Shām / Evening in Swat)
}
& NUT & 0.2165 & 3.0000 \\

4 &
\makecell[l]{
\foreignlanguage{arabic}{آخری بس} \\
(Ākhrī Bas / The Last Bus)
}
& NUT & 0.3025 & 5.0000 \\

5 &
\makecell[l]{
\foreignlanguage{arabic}{عید کا نیا چاند} \\
(ʿĪd kā Nayā Chānd / The New Moon of Eid)
}
& NUT & 0.1254 & 2.0000 \\

6 &
\makecell[l]{
\foreignlanguage{arabic}{ایمان کی تلاش} \\
(Īmān kī Talāsh / Search for Faith)
}
& NUT & 0.2215 & 3.3333 \\

7 &
\makecell[l]{
\foreignlanguage{arabic}{پیلا وال پیپر} \\
(Pīlā Vāl Pēpar / The Yellow Wallpaper)
}
& TSST & 0.3524 & 4.0000 \\

8 &
\makecell[l]{
\foreignlanguage{arabic}{لاٹری} \\
(Lāṭrī / The Lottery)
}
& TSST & 0.9110 & 15.0000 \\

9 &
\makecell[l]{
\foreignlanguage{arabic}{بندر کا پنجہ} \\
(Bandar kā Panjah / The Monkey's Paw)
}
& TSST & 0.3188 & 5.6667 \\

10 &
\makecell[l]{
\foreignlanguage{arabic}{دوسری عورت} \\
(Dūsrī ʿAurat / The Second Woman)
}
& UST & 0.2804 & 5.6667 \\

11 &
\makecell[l]{
\foreignlanguage{arabic}{ٹھنڈا گوشت} \\
(Ṭhanḍā Gōsht / Cold Meat)
}
& UST & 0.3720 & 5.6667 \\

12 &
\makecell[l]{
\foreignlanguage{arabic}{خاموش اذان} \\
(Khāmōsh Az̤ān / Silent Call to Prayer)
}
& UST & 0.4921 & 7.3333 \\

13 &
\makecell[l]{
\foreignlanguage{arabic}{خرگوش اور کچھوا} \\
(Khargōsh aur Kachhwā / The Hare and the Tortoise)
}
& Fable & 0.0219 & 0.3333 \\

14 &
\makecell[l]{
\foreignlanguage{arabic}{پیاسا کوا} \\
(Piyāsā Kawwā / The Thirsty Crow)
}
& Fable & 0.0228 & 0.3333 \\

15 &
\makecell[l]{
\foreignlanguage{arabic}{بیوقوف گدھا} \\
(Bēwaqūf Gadhā / The Foolish Donkey)
}
& Fable & 0.1589 & 3.0000 \\

16 &
\makecell[l]{
\foreignlanguage{arabic}{اندھا آدمی اور ہاتھی} \\
(Andhā Ādmī aur Hāthī / The Blind Man and the Elephant)
}
& TSST & 0.0000 & 0.0000 \\

17 &
\makecell[l]{
\foreignlanguage{arabic}{لحاف} \\
(Liḥāf / The Quilt)
}
& UST & 1.7525 & 26.6667 \\

18 &
\makecell[l]{
\foreignlanguage{arabic}{جلا وطن} \\
(Jalā Waṭan / Exiled)
}
& UST & 0.2507 & 4.3333 \\

19 &
\makecell[l]{
\foreignlanguage{arabic}{آگ کا دریا} \\
(Āg kā Daryā / River of Fire)
}
& UST & 0.3410 & 5.0000 \\

20 &
\makecell[l]{
\foreignlanguage{arabic}{گرم کوٹ} \\
(Garm Koṭ / Warm Coat)
}
& UST & 0.7702 & 10.6667 \\

21 &
\makecell[l]{
\foreignlanguage{arabic}{اوور کوٹ} \\
(Ōvar Koṭ / The Overcoat)
}
& UST & 1.2524 & 21.3333 \\

22 &
\makecell[l]{
\foreignlanguage{arabic}{لالچی کتا} \\
(Lālchī Kuttā / The Greedy Dog)
}
& Fable & 0.0184 & 0.3333 \\

23 &
\makecell[l]{
\foreignlanguage{arabic}{پچھتاوا} \\
(Pachtāwā / Regret)
}
& TSST & 0.2198 & 4.0000 \\

24 &
\makecell[l]{
\foreignlanguage{arabic}{سر کٹا سوار} \\
(Sar Kaṭā Sawār / The Headless Horseman)
}
& TSST & 0.3038 & 5.3333 \\

25 &
\makecell[l]{
\foreignlanguage{arabic}{ایک گھنٹے کی کہانی} \\
(Ek Ghanṭē kī Kahānī / The Story of an Hour)
}
& TSST & 0.0791 & 1.3333 \\

26 &
\makecell[l]{
\foreignlanguage{arabic}{پہاڑی کی برف} \\
(Pahāṛī kī Barf / Snow of the Mountain)
}
& TSST & 0.1813 & 3.0000 \\

27 &
\makecell[l]{
\foreignlanguage{arabic}{لڑکے اور لڑکیاں} \\
(Laṛkē aur Laṛkiyāṅ / Boys and Girls)
}
& TSST & 0.1513 & 2.3333 \\

28 &
\makecell[l]{
\foreignlanguage{arabic}{سلطانہ کا خواب} \\
(Sulṭānah kā Khwāb / Sultana's Dream)
}
& UST & 0.0156 & 0.3333 \\

29 &
\makecell[l]{
\foreignlanguage{arabic}{خاتون یا شیر} \\
(Khātūn yā Shēr / The Lady or the Tiger)
}
& TSST & 0.0755 & 1.3333 \\

30 &
\makecell[l]{
\foreignlanguage{arabic}{خط} \\
(Khaṭ / The Letter)
}
& TSST & 2.3459 & 36.6667 \\

31 &
\makecell[l]{
\foreignlanguage{arabic}{پت جھڑ کی آواز} \\
(Pat Jhaṛ kī Āwāz / Sound of Autumn Leaves)
}
& UST & 0.2860 & 5.0000 \\

\hline
\end{tabular}
\end{table*}
\section{Script Contamination: Full Analysis}
\label{app:script_contamination}

\subsection{Detection Method}
\label{app:script_contamination_method}

A word was flagged as script-contaminated when any
character in it fell outside the Urdu Unicode range
(U+0600 to U+06FF) and the standard Arabic presentation
forms (U+FB50 to U+FDFF, U+FE70 to U+FEFF), excluding
digits, punctuation and whitespace. We classify each
contaminated word into two structural patterns:

\begin{itemize}
  \item \textbf{Pure intrusion.} The entire word belongs
        to a foreign script (e.g.\ \textit{hospital},
        \textit{career}, \textit{water}).
  \item \textbf{Partial contamination.} The word mixes
        characters from Urdu and at least one foreign
        script (e.g.\
        \foreignlanguage{arabic}{ح}aj\foreignlanguage{arabic}{ی}
        for \foreignlanguage{arabic}{حاجی}).
\end{itemize}

Aggregate counts per script and per model are given in
Table~\ref{tab:script_contamination}.

\subsection{Latin Script Intrusion}
\label{app:contamination_latin}

Latin was the most frequent contaminant across all three
models. Pure English-word substitution included everyday
vocabulary for which standard Urdu equivalents exist:
\textit{hospital}, \textit{career}, \textit{water},
\textit{warmth}, \textit{tight}.

Two partial-contamination cases warrant additional
discussion beyond what the main text reports. First, QWEN's
substitution of
\foreignlanguage{arabic}{حاجی} with
\foreignlanguage{arabic}{ح}aj\foreignlanguage{arabic}{ی}
replaces the vowel graphemes
\foreignlanguage{arabic}{ا} and \foreignlanguage{arabic}{ی}
with their Latin phonetic equivalents \textit{a} and
\textit{j}. The form recurs 10 times in the same
1{,}343-word story and never spaces out the Latin
characters, indicating that the token boundary itself is
broken in the model's vocabulary. Second, DeepSeek
produced \foreignlanguage{arabic}{کہ}a, replacing the
final \foreignlanguage{arabic}{ا} of
\foreignlanguage{arabic}{کہا} (\textit{said}) with Latin
\textit{a}: phonetically equivalent but orthographically
invalid. The substitution suggests that the decoder
resolved an ambiguous token boundary by falling back to
Latin. DeepSeek produced
\foreignlanguage{arabic}{رسمی}ities, attaching the
English plural morpheme \textit{-ities} to the Urdu
adjective \foreignlanguage{arabic}{رسمی}
(\textit{formal}), a previously undocumented form of
cross-lingual morphological contamination in which English
derivational morphology is applied to an Urdu stem.

\subsection{Devanagari (Hindi) Intrusion}
\label{app:contamination_devanagari}

Devanagari intrusion was most prevalent in DeepSeek and
also appeared in GPT and QWEN. Because Urdu and Hindi
share extensive phonological overlap, the proximity
produces character-level cross-script leakage in which
phonetically equivalent Devanagari diacritics substitute
for their Urdu counterparts. For example,
\foreignlanguage{arabic}{ٹھ}\foreignlanguage{hindi}{ं}\foreignlanguage{arabic}{ڈی}
(\textit{cold}) carries an internal Devanagari nasaliser.
More severe is
\foreignlanguage{arabic}{ک}\foreignlanguage{hindi}{ुत्त}\foreignlanguage{arabic}{وں},
where the Urdu word for \textit{dogs} is partially
rendered in Devanagari mid-token: a morpheme-level script
switch rather than a diacritic substitution.

\subsection{Chinese and Cyrillic Intrusion}
\label{app:contamination_cjk_cyrillic}

The most striking contamination instances involve Chinese
(CJK) and Cyrillic characters embedded within Urdu tokens.
These scripts have low or no phonological or typological
relationship to Urdu, making their presence entirely
anomalous. In every observed case the foreign token is
semantically appropriate in context: \textit{yánsù}
(\textit{serious}), \textit{de shēnghuó}
(\textit{of life}), \textit{huàtí} (\textit{topic}),
\textit{iskrennost'} (\textit{sincerity}). The failure is
orthographic rather than semantic: DeepSeek selected a
valid lexical item from its Chinese or Cyrillic
vocabulary where an Urdu equivalent was required, and
omitted the delimiting space, producing undelimited
bilingual compounds. The exception is
\foreignlanguage{arabic}{پراج}\foreignlanguage{russian}{ек}\foreignlanguage{arabic}{ٹ},
where Cyrillic characters are substituted for similar Urdu
phonemes within a single word, a pattern structurally
analogous to the Devanagari substitutions above.
\begin{table}[ht]

\centering
\caption{Script Contamination Summary by Model}
\label{tab:script_contamination}
\scalebox{0.7}{
\begin{tabular}{lcccc}
\hline
\textbf{Metric} & 
\textbf{GPT} & 
\textbf{QWEN} & 
\textbf{DeepSeek} & 
\textbf{Total} \\
\hline
Affected stories        & 6  & 3  & 15 & 24 \\
Total contaminated words& 11 & 14 & 61 & 86 \\
\quad Pure intrusion    & 8  & 3  & 42 & 53 \\
\quad Partial mixed     & 3  & 11 & 19 & 33 \\
\hline
\multicolumn{5}{l}{\textit{By Script}} \\
\quad Latin (English)   & 7  & 13 & 47 & 67 \\
\quad Devanagari (Hindi)& 3  & 1  & 9  & 13 \\
\quad Chinese (CJK)     & 0  & 0  & 3  & 3  \\
\quad Cyrillic          & 1  & 0  & 2  & 3  \\
\hline
\end{tabular}
}
\end{table}

\section{Top Repetition Cases}
\label{app:top_repetition}

Table~\ref{tab:top_repetition_cases} lists individual
(title, model) pairs with the highest absolute repetition
counts. UST titles dominate the list, consistent with the
Lexical Title Unpacking interpretation.

\begin{table*}[h]
\centering\scriptsize
\caption{Top repetition cases across literary title
sources and LLMs. TSST = Translated Short Story Title;
UST = Urdu Short Story Title.}
\label{tab:top_repetition_cases}
\renewcommand{\arraystretch}{1.2}
\begin{tabular}{p{7cm} c l c}
\hline
\textbf{Title (ALA-LC / English)} &
\textbf{Source} & \textbf{Model} &
\textbf{Count (Rate per 100)} \\
\hline
\multirow{2}{*}{\makecell[l]{
\foreignlanguage{arabic}{خط} (Khaṭ / The Letter)}}
  & \multirow{2}{*}{TSST}
  & DeepSeek & 61 (3.27) \\
  &  & GPT & 33 (2.08) \\
\hline
\multirow{3}{*}{\makecell[l]{
\foreignlanguage{arabic}{لحاف} (Liḥāf / The Quilt)}}
  & \multirow{3}{*}{UST}
  & QWEN     & 35 (2.79) \\
  &  & DeepSeek & 28 (1.30) \\
  &  & GPT      & 17 (1.17) \\
\hline
\multirow{2}{*}{\makecell[l]{
\foreignlanguage{arabic}{اوور کوٹ} (Ōvar Koṭ / The Overcoat)}}
  & \multirow{2}{*}{UST}
  & DeepSeek & 35 (1.76) \\
  &  & QWEN     & 19 (1.41) \\
\hline
\makecell[l]{
\foreignlanguage{arabic}{لاٹری} (Lāṭrī / The Lottery)}
  & TSST & GPT & 20 (1.16) \\
\hline
\makecell[l]{
\foreignlanguage{arabic}{گرم کوٹ} (Garm Koṭ / Warm Coat)}
  & UST  & DeepSeek & 17 (1.03) \\
\hline
\makecell[l]{
\foreignlanguage{arabic}{خاموش اذان} (Khāmōsh Az̤ān / Silent Call to Prayer)}
  & UST  & GPT & 16 (1.14) \\
\hline
\end{tabular}
\end{table*}

\subsection{Original-Author vs LLM TRR}
\label{app:trr_original}

Table~\ref{tab:trr_original} compares LLM TRR against the
original-author baseline for the four canonical Urdu
titles whose source texts were obtained from Rekhta.

\begin{table*}[h]
\centering\small
\caption{TRR: original authors vs LLM-generated stories.}
\label{tab:trr_original}
\begin{tabular}{llccccc}
\hline
\textbf{Title} & \textbf{Author} &
\textbf{\shortstack{Orig\\TRR}} &
\textbf{GPT} & \textbf{QWEN} & \textbf{DS} &
\textbf{\shortstack{Avg\\Ratio}} \\
\hline
\foreignlanguage{arabic}{لحاف}      & Ismat Chughtai
  & 0.52 & 1.17 {\small(2.2$\times$)}
  & 2.79 {\small(5.3$\times$)}
  & 1.30 {\small(2.5$\times$)} & 3.3$\times$ \\
\foreignlanguage{arabic}{اوور کوٹ}  & Ghulam Abbas
  & 0.41 & 0.59 {\small(1.4$\times$)}
  & 1.41 {\small(3.4$\times$)}
  & 1.76 {\small(4.3$\times$)} & 3.0$\times$ \\
\foreignlanguage{arabic}{گرم کوٹ}   & Rajinder Singh Bedi
  & 0.33 & 0.16 {\small(0.5$\times$)}$^\dagger$
  & 1.13 {\small(3.4$\times$)}
  & 1.03 {\small(3.1$\times$)} & 2.3$\times$ \\
\foreignlanguage{arabic}{ٹھنڈا گوشت} & Saadat Hasan Manto
  & 0.04 & 0.48 {\small(10.6$\times$)}
  & 0.05 {\small(1.3$\times$)}
  & 0.57 {\small(12.7$\times$)} & 8.2$\times$ \\
\hline
\multicolumn{7}{l}{{\small $^\dagger$ GPT decomposed the
title, using \foreignlanguage{arabic}{کوٹ} alone
extensively (see Section~\ref{subsec:trr_limit}).}}\\
\hline
\end{tabular}
\end{table*}

\section{Memorisation Analysis: Method and Full Results}
\label{app:memorisation}

\subsection{Method}
\label{app:memorisation_method}

For each \emph{(original, generated)} story pair
($n=36$ pairs in total: 12 titles $\times$ 3 models) we
asked GPT-4o-mini at temperature~0 with JSON response
format to produce four overlap scores in $[0,1]$ and a
categorical label.

\paragraph{Axes.}
\begin{itemize}
  \item \textbf{Plot overlap.} The judge first extracts
        4 to 8 character-agnostic plot beats from the
        original (e.g.\ ``a thirsty animal finds water it
        cannot reach''), then reports the fraction of
        those beats that recur in the generated story in
        the same causal order.
  \item \textbf{Theme overlap.} Semantic similarity of the
        central moral or message of the two stories.
  \item \textbf{Character overlap.} Set-style overlap of
        the principal characters and their narrative
        functions, insensitive to renaming.
  \item \textbf{Resolution match.} Semantic similarity of
        the one-sentence narrative endings, compared in
        isolation. The judge also returns a categorical
        verdict
        $\in\{\texttt{same},\texttt{partial},\texttt{different}\}$.
\end{itemize}

\paragraph{Thresholds.}
Each axis is converted to a boolean using a fixed
threshold: plot~$\geq$~0.60, theme~$\geq$~0.70,
resolution~$\geq$~0.80. A generated story is labelled
\emph{memorisation} when at least two of these three
booleans are true. Crossed with character overlap
(threshold 0.50), this yields the 2$\times$2 typology of
Table~\ref{tab:memorisation_typology}.

{These thresholds are corpus-specific, single-annotator
operating points rather than values drawn from a prior
theoretical framework. They were set by manual calibration: a
native Urdu-speaking researcher compared each generated story
with its original and judged whether a }\emph{{Faithful
Retelling}} {label was appropriate; the cut-offs above are the
values that best matched those judgements. Plot received a
lower threshold than theme or resolution because it is
evaluated across multiple narrative beats, whereas theme and
resolution each require a single, stronger judgement of
semantic equivalence.
}

end \paragraph{Categorical label.}
Independent of the threshold rule, the judge also assigns
one of four discrete labels:
\textbf{PR} (Plot Reproduction: plot, theme, characters
and setting all substantially preserved);
\textbf{PI} (Partial Inheritance: two or three axes
partially overlap);
\textbf{OR} (Original: only the title or a high-level
topic is shared);
\textbf{LI} (Lexical Inheritance: title and surface motifs
reused but no narratological axis aligns; no story in our
corpus received this label).

\begin{table}[h]
\centering\small
\caption{Memorisation typology (definitions).}
\label{tab:memorisation_typology}
\begin{tabular}{lcc}
\hline
                            & \textbf{Char.\ LOW} & \textbf{Char.\ HIGH} \\
\hline
\textbf{$\geq 2$ axes agree} & Disguised Mem.\ & Faithful Retelling \\
\textbf{$< 2$ axes agree}    & Original         & Spin-off \\
\hline
\end{tabular}
\end{table}

\subsection{Per-Category Breakdown}
\label{app:memorisation_breakdown}

Table~\ref{tab:mem_by_category} disaggregates outcomes by
category. Fables dominate the \emph{Faithful Retelling}
cell for GPT and DeepSeek; English classics produce the
mixed pattern (one Chopin PR for DeepSeek, all others
\emph{Original}); every Urdu-canon row is \emph{Original}
for every model.

\begin{table}[h]
\centering\small
\caption{Memorisation label counts by category and model.
PR = Plot Reproduction; PI = Partial Inheritance;
OR = Original.}
\label{tab:mem_by_category}
\scalebox{0.77}{
\begin{tabular}{lccc|ccc|ccc}
\hline
& \multicolumn{3}{c|}{\textbf{Fable ($n=4$)}}
& \multicolumn{3}{c|}{\textbf{English ($n=3$)}}
& \multicolumn{3}{c}{\textbf{Urdu canon ($n=5$)}} \\
\textbf{Model} & PR & PI & OR & PR & PI & OR & PR & PI & OR \\
\hline
GPT       & 3 & 0 & 1 & 0 & 0 & 3 & 0 & 1 & 4 \\
DeepSeek  & 2 & 1 & 1 & 1 & 0 & 2 & 0 & 0 & 5 \\
QWEN      & 1 & 0 & 3 & 0 & 1 & 2 & 0 & 0 & 5 \\
\hline
\end{tabular}
}
\end{table}

\subsection{Cross-Model Plot Overlap}
\label{app:memorisation_cross_model}

For each title we also computed pairwise overlap between
the three generated versions
(Table~\ref{tab:mem_cross_model}). GPT and DeepSeek
overlap most often (mean 0.45), consistent with both
models tending toward Faithful Retelling on the same
fables. QWEN is roughly equidistant from both (0.33 to
0.34) and contributes new plots more often than it
mirrors either.

\begin{table}[h]
\centering\small
\caption{Mean pairwise overlap between the three generated
versions of the same source title ($n=12$ titles).}
\label{tab:mem_cross_model}
\begin{tabular}{lcc}
\hline
\textbf{Model pair}  & \textbf{Mean} & \textbf{Median} \\
\hline
GPT \& DeepSeek       & 0.45 & 0.35 \\
DeepSeek \& QWEN      & 0.34 & 0.28 \\
GPT \& QWEN           & 0.33 & 0.35 \\
\hline
\end{tabular}
\end{table}

\section{Cross-Model Structural Analysis: Details}
\label{app:network_details}

\subsection{Extraction Protocol}
\label{app:network_extraction}

For each of the 93 LLM-generated stories, the following
instruction was issued to GPT-5.4 at temperature 0 with
JSON response format enforced. The judge was given the
story title and the full story text (truncated at
18{,}000 characters, a limit no generated story
exceeded).

\begin{quote}\small\itshape
You are a literary analyst fluent in Urdu. Extract:
(1)~the list of distinct characters who actually appear
or speak in the story (skip purely mentioned-by-name
characters with no role); (2)~the signed pairwise
relationships between those characters as evidenced by
the story. For each character record a name and a role
(protagonist, antagonist, supporting, mentor,
love-interest, antagonised-victim, narrator, other).
For each relationship record source, target, sign
(\textit{positive} = affection / support / alliance /
love / friendship; \textit{negative} = conflict /
hostility / betrayal / abuse / rivalry;
\textit{neutral} = acquaintance / professional /
functional), intensity $\in \{1,2,3\}$, and a
$\leq$20-word Urdu evidence paraphrase. Only include
relationships with textual support; use the same
spelling of a name across all of that character's
edges; do not invent characters; list symmetric
relationships once.
\end{quote}

\subsection{Metric Definitions}
\label{app:network_metrics}

Given a story-level signed weighted graph $G=(V,E)$ with
edge weight
$w_e = \text{sign}(e)\cdot\text{intensity}(e) \in
\{-3,\ldots,+3\}$:

\begin{itemize}
  \item \textbf{Density} $= 2|E|/(|V|(|V|-1))$.
  \item \textbf{Average clustering}: mean unweighted
        local clustering coefficient over all nodes.
  \item \textbf{Positivity ratio} $= |\{e : w_e > 0\}| /
        |\{e : w_e \neq 0\}|$ (positive share of
        \emph{signed} edges; neutral edges excluded).
  \item \textbf{Mean signed edge weight} $=
        \frac{1}{3|E|}\sum_{e \in E} w_e$, rescaled to
        $[-1,+1]$.
  \item \textbf{Jaccard character overlap} between
        models $A$ and $B$ on the same prompt $=
        |C_A \cap C_B|/|C_A \cup C_B|$, where $C_X$ is
        the set of normalised character names produced
        by model $X$.
\end{itemize}

Duplicate edges between the same character pair are
averaged before any metric is computed.

\subsection{Per-Story Distributions}
\label{app:network_distributions}

Figures~\ref{fig:density_violin} and 
\ref{fig:intensity_violin} 
 show per-story distributions
of density and mean signed edge weight.

\begin{figure}[h]
\centering
\includegraphics[width=0.85\linewidth]{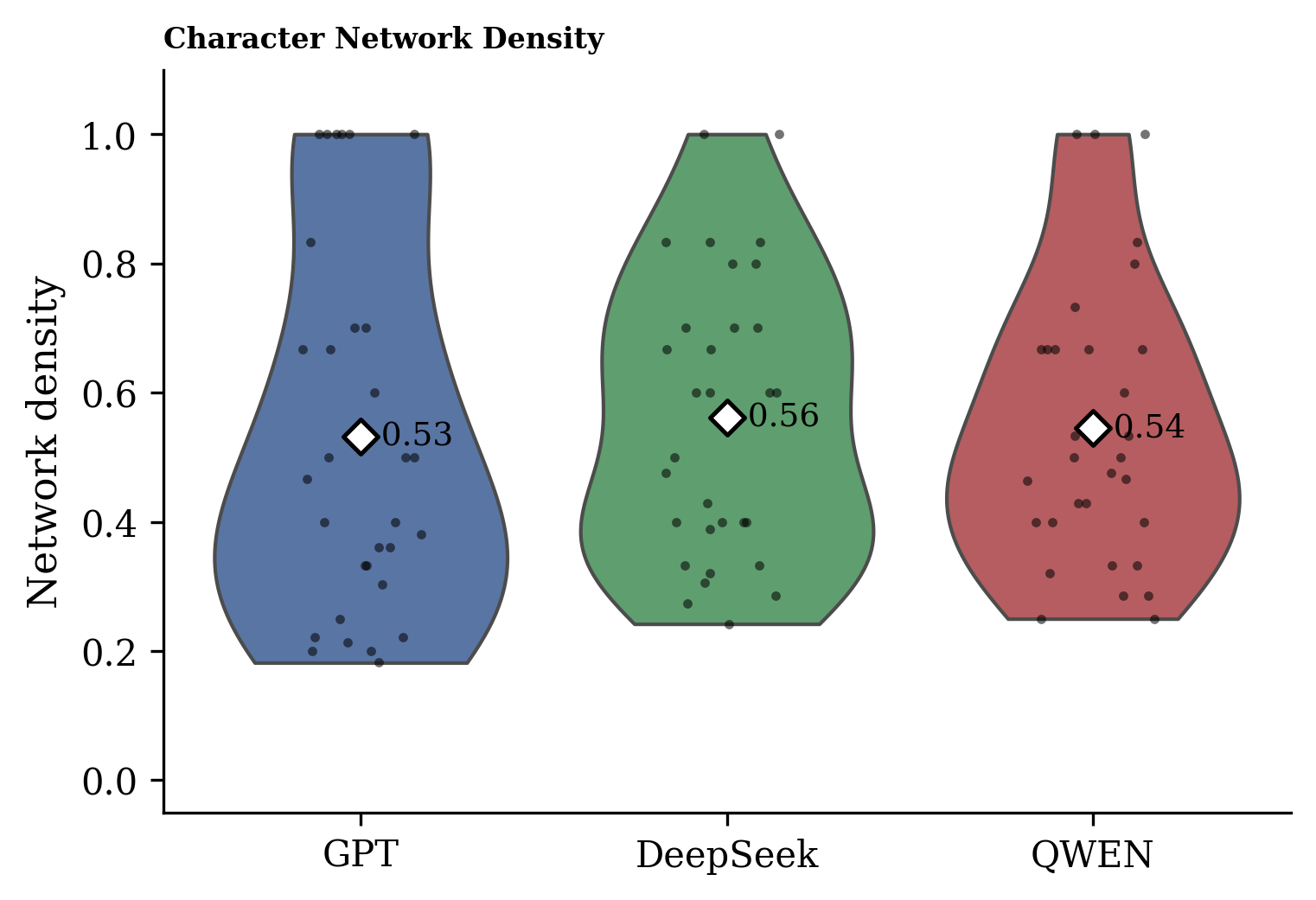}
\caption{Density per story. White diamonds mark means.}
\label{fig:density_violin}
\end{figure}

\begin{figure}[h]
\centering
\includegraphics[width=0.85\linewidth]{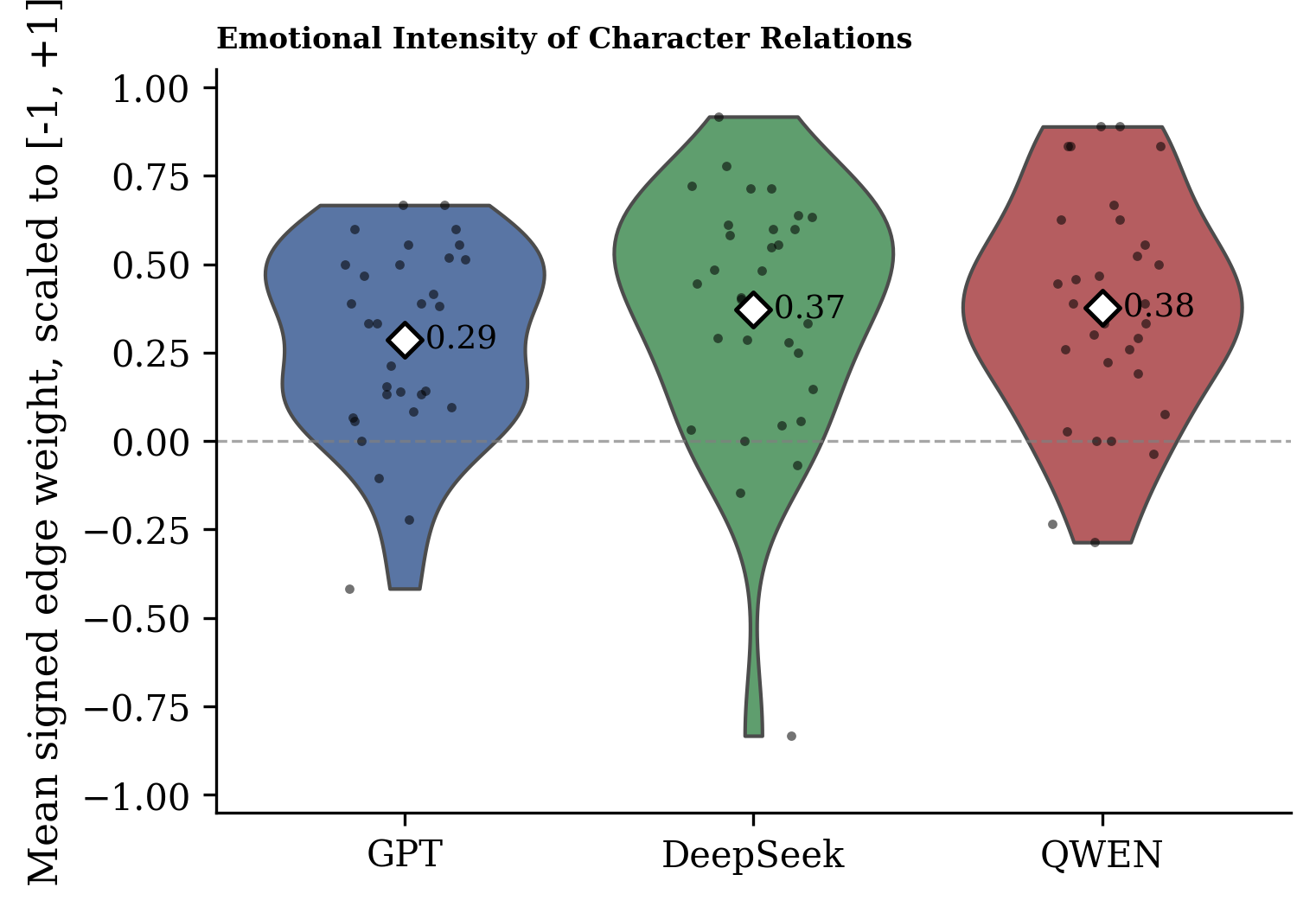}
\caption{Mean signed edge weight per story on $[-1,+1]$.}
\label{fig:intensity_violin}
\end{figure}

\subsection{Why Human Stories Are Excluded}
\label{app:human_exclusion}

The human reference corpus consists of published Urdu
short fiction by canonical authors (Manto, Chughtai,
Bedi, Krishan Chander). These stories run
5{,}000 to 15{,}000 words against an LLM target of
2{,}000. A longer story has room for incidental and
functional characters (shopkeepers, neighbours, passing
acquaintances) that appear in the network as additional
neutral edges, inflating character counts and the
neutral edge share. A direct human-vs-LLM comparison on
these metrics would therefore be confounded by length.
Truncating the human stories to 2{,}000 words was
considered and rejected, as it would discard the
resolution and closing character introductions of most
published short stories and replace a length confound
with a worse truncation artefact. The network analysis
is therefore restricted to the three LLM corpora, with a
length-matched human comparison left to future work.
begin

\section{{Annotation: Per-Model and Per-Label Agreement}}
\label{app:iaa}

{Table~\ref{tab:iaa_by_model} reports Cohen's $\kappa$ computed
separately for the stories generated by each model, on the
60-story inter-annotator-agreement subset described in
Section~\ref{subsec:annotation}. Agreement is consistently
substantial and does not vary meaningfully by source model.
}

\begin{table}[h]
\centering\small
\caption{{Inter-annotator agreement by model (60-story IAA
subset).}}
\label{tab:iaa_by_model}
\begin{tabular}{lcc}
\hline
\textbf{{Model}} & \textbf{{Cohen's $\kappa$}} & \textbf{{Interpretation}} \\
\hline
{GPT      }& {0.8072 }& {Substantial }\\
{DeepSeek }& {0.8084 }& {Substantial }\\
{QWEN     }& {0.8058 }& {Substantial }\\
\hline
\textbf{{Overall}} & \textbf{{0.81}} & \textbf{{Substantial}} \\
\hline
\end{tabular}
\end{table}

{We do not report agreement broken down by individual label,
because the number of instances of several labels (e.g.\
Cultural Inconsistency) in the 60-story subset is too small to
support a meaningful per-label estimate.
}

\section{{Repeated Generations: Full Results}}
\label{app:robustness_full}

{To estimate within-condition variance, five titles (IDs 6, 8,
10, 13 and 18, one per title category, selected with a fixed
random seed) were regenerated twice more per model, using the
same prompt as the main corpus, yielding 30 additional Urdu
stories.
}

\begin{table}[h]
\centering\small
\caption{{Title Repetition Rate across three independent runs
(5 titles).}}
\label{tab:robust_trr}
\begin{tabular}{lccccc}
\hline
\textbf{{Model}} & \textbf{{R1}} & \textbf{{R2}} & \textbf{{R3}} & \textbf{{Mean}} & \textbf{{SD}} \\
\hline
{GPT      }& {0.100 }& {0.140 }& {0.218 }& {0.153 }& {0.048 }\\
{QWEN     }& {0.258 }& {0.219 }& {0.193 }& {0.223 }& {0.027 }\\
{DeepSeek }& {0.170 }& {0.150 }& {0.276 }& {0.199 }& {0.055 }\\
\hline
\end{tabular}
\end{table}

\begin{table}[h]
\centering\small
\caption{{Raw title-occurrence count per story across three runs
(5 titles).}}
\label{tab:robust_count}
\begin{tabular}{lccccc}
\hline
\textbf{{Model}} & \textbf{{R1}} & \textbf{{R2}} & \textbf{{R3}} & \textbf{{Mean}} & \textbf{{SD}} \\
\hline
{GPT      }& {1.80 }& {2.80 }& {4.40 }& {3.00 }& {1.06 }\\
{QWEN     }& {4.20 }& {3.00 }& {2.80 }& {3.33 }& {0.60 }\\
{DeepSeek }& {2.80 }& {3.00 }& {5.20 }& {3.67 }& {1.08 }\\
\hline
\end{tabular}
\end{table}

{TRR's SD ranges from 0.027 to 0.055 across models, indicating
moderate run-to-run variation in magnitude. The direction of the
Lexical Title Unpacking effect is nonetheless stable: full-corpus
Urdu-canon TRR (0.40--0.43 across models) exceeds the
re-generated TRR in all nine model$\times$run comparisons. Raw
counts show the same pattern independently of story length,
with Urdu mean counts (5.19--8.12) exceeding re-generated means
(3.00--3.67) across all models and runs.
}

\begin{table}[h]
\centering\small
\caption{{Positivity ratio in character networks across three
runs (5 titles).}}
\label{tab:robust_positivity}
\begin{tabular}{lccccc}
\hline
\textbf{{Model}} & \textbf{{R1}} & \textbf{{R2}} & \textbf{{R3}} & \textbf{{Mean}} & \textbf{{SD}} \\
\hline
{GPT      }& {0.780 }& {0.803 }& {0.858 }& {0.814 }& {0.032 }\\
{DeepSeek }& {0.820 }& {0.880 }& {0.778 }& {0.826 }& {0.042 }\\
{QWEN     }& {0.820 }& {0.719 }& {0.820 }& {0.786 }& {0.048 }\\
\hline
\end{tabular}
\end{table}

{Positivity ratio is the most stable finding of the three: SD
ranges from 0.032 to 0.048, and all nine values (3 models
$\times$ 3 runs) exceed 0.71. The positivity bias documented in
Section~\ref{sec:networks} is therefore not an artefact of a
single stochastic generation.
}

\begin{table}[h]
\centering\small
\caption{{Protagonist gender across three runs (5 titles).}}
\label{tab:robust_gender}
\begin{tabular}{lcccc}
\hline
\textbf{{Model}} & \textbf{{R1 M:F}} & \textbf{{R2 M:F}} & \textbf{{R3 M:F}} & \textbf{{Pooled M:F}} \\
\hline
{GPT      }& {4:1 }& {3:1 }& {3:1 }& {10:3 (3.3:1) }\\
{DeepSeek }& {4:0 }& {3:1 }& {2:1 }& {9:2 (4.5:1) }\\
{QWEN     }& {5:0 }& {3:1 }& {2:1 }& {10:2 (5.0:1) }\\
\hline
\multicolumn{4}{l}{\textbf{Pooled overall}} & \textbf{{29:7 (4.1:1)}} \\
\hline
\end{tabular}
\end{table}

{Protagonist gender distribution is stable across all three runs:
male protagonists outnumber female protagonists in every run and
every model, with no run producing female-majority output. The
pooled ratio (4.1:1) is consistent with the full-corpus finding
(3.7:1).
}

\begin{table*}[h]
\centering\small
\caption{{Manual annotation counts on the two additional runs
(R1, R2), by error type and model.}}
\label{tab:robust_annotation}
\begin{tabular}{lcccccc}
\hline
\textbf{{Model}} & \textbf{{Ling.\ R1}} & \textbf{{Ling.\ R2}} & \textbf{{Sem.\ R1}} & \textbf{{Sem.\ R2}} & \textbf{{Tot.\ R1}} & \textbf{{Tot.\ R2}} \\
\hline
{GPT      }& {15 }& {35 }& {4  }& {12 }& {19 }& {47 }\\
{DeepSeek }& {35 }& {15 }& {11 }& {2  }& {46 }& {17 }\\
{QWEN     }& {18 }& {11 }& {6  }& {8  }& {24 }& {19 }\\
\hline
\textbf{{Total}} & \textbf{{68}} & \textbf{{61}} & \textbf{{21}} & \textbf{{22}} & \textbf{{89}} & \textbf{{83}} \\
\hline
\end{tabular}
\end{table*}

{Manual annotation of the two additional runs confirms that
linguistic and semantic errors persist across independent
generations: errors are present in every run for every model,
and no model produces error-free output in either run. Recurring
error types mirror those documented in the main corpus. For
agreement errors, QWEN produced
}\foreignlanguage{arabic}{موبائل فونوں کا چھوٹا سا شاپ} {(``a
small mobile phone shop''), where the masculine noun ``shop''
requires a masculine modifier but the broader noun phrase is
malformed. For semantic anomaly, a DeepSeek sentence introduced a
character as ``He was Haji Fazl, the one with soggy bread'',
syntactically well-formed but semantically incoherent as an
identifying description. For context mismatch, a mother
addressing her son said
}\foreignlanguage{arabic}{تو تو میرے سامنے بیٹھا ہے، میرے ابّا کی
دعا قبول ہو گئی} {(``You are sitting before me; my father's
prayer was answered''), where ``my father'' refers to the
mother's own father rather than the son's, a kinship error.
Cultural and script-level errors also recurred: GPT mixed
Punjabi phrases into an otherwise Urdu narrative and embedded
Latin-script English words within Urdu text; QWEN's retelling of
}\emph{{The Hare and the Tortoise}} {introduced a Christmas
card-making competition, a Western cultural substitution with no
grounding in Pakistani context; and em-dash usage recurred in GPT
stories, consistent with the stylistic-transfer finding of
Section~\ref{subsec:auto_errors}.
}

\section{{English-Language Control: Full Results}}
\label{app:english_control}

{To separate what is specific to Urdu generation from what is a
general property of these three models, the same three models
were prompted with the same seven TSST titles in English (21
stories), evaluated on the three dimensions directly comparable
across languages: positivity bias, memorisation behaviour, and
title repetition rate. Script contamination, space deletion, and
em-dash usage were not carried over, as they are either specific
to Arabic-script generation or normatively acceptable in English
prose.
}

\begin{table}[h]
\centering\small
\caption{{Character-network sentiment: Urdu vs.\ English.}}
\label{tab:english_positivity}
\begin{tabular}{lcccc}
\hline
\textbf{{Model}} & \textbf{{Ur.\ Pos.}} & \textbf{{En.\ Pos.}} & \textbf{{Ur.\ Edge}} & \textbf{{En.\ Edge}} \\
\hline
{GPT      }& {0.78 }& {0.74 }& {0.29 }& {0.18 }\\
{QWEN     }& {0.82 }& {0.60 }& {0.38 }& {0.15 }\\
{DeepSeek }& {0.82 }& {0.57 }& {0.37 }& {0.00 }\\
\hline
\textbf{{Overall}} & \textbf{{0.807}} & \textbf{{0.637}} & \textbf{{0.347}} & \textbf{{0.110}} \\
\hline
\end{tabular}
\end{table}

{The positivity ratio is 17 points higher in Urdu (0.807) than in
English (0.637); the gap is 22 and 25 points for QWEN and
DeepSeek respectively, while GPT shows the smallest difference
(0.78 vs.\ 0.74), suggesting its positivity bias is less
language-dependent. Mean signed edge weight follows the same
pattern: DeepSeek scores 0.00 in English versus 0.37 in Urdu,
indicating that emotional amplification in Urdu stories is more
intense as well as more positive.
}

\begin{table*}[h]
\centering\small
\caption{{Memorisation outcomes: Urdu vs.\ English (7 TSST titles).}}
\label{tab:english_memorisation}
\begin{tabular}{lcccc}
\hline
\textbf{{Model}} & \textbf{{Ur.\ FR\%}} & \textbf{{En.\ FR\%}} & \textbf{{Ur.\ Comp.}} & \textbf{{En.\ Comp.}} \\
\hline
{GPT      }& {25.0 }& {14.3 }& {0.50 }& {0.386 }\\
{QWEN     }& \phantom{0}{8.3 }& {42.9 }& {0.27 }& {0.599 }\\
{DeepSeek }& {25.0 }& {57.1 }& {0.43 }& {0.664 }\\
\hline
\textbf{{Overall}} & \textbf{{19.4}} & \textbf{{38.1}} & \textbf{{0.40}} & \textbf{{0.550}} \\
\hline
\end{tabular}
\end{table*}

\begin{table}[h]
\centering\small
\caption{{Cross-model plot overlap (mean Jaccard): Urdu vs.\ English.}}
\label{tab:english_crossmodel}
\begin{tabular}{lcc}
\hline
\textbf{{Model pair}} & \textbf{{Urdu}} & \textbf{{English}} \\
\hline
{GPT -- DeepSeek }& {0.45 }& {0.393 }\\
{GPT -- QWEN     }& {0.33 }& {0.486 }\\
{DeepSeek -- QWEN }& {0.34 }& {0.671 }\\
\hline
\textbf{{Overall}} & \textbf{{0.37}} & \textbf{{0.517}} \\
\hline
\end{tabular}
\end{table}

{English stories show nearly double the faithful-retelling rate
of Urdu (38.1\% vs.\ 19.4\%), and cross-model plot overlap is
higher in English (0.517 vs.\ 0.373). Both findings are
consistent with stronger training-data coverage of English
source texts. Model rankings reverse across languages: QWEN is
the weakest memoriser in Urdu (8.3\%) but second-highest in
English (42.9\%), and DeepSeek is strongest in English (57.1\%)
but middle-ranked in Urdu (25.0\%) -- each model's memorisation
behaviour appears to reflect its own language-specific training
coverage rather than a fixed model property. The higher
}\emph{{Original}} {label rate in Urdu does not reflect creativity;
as the TRR analysis shows, it reflects the absence of
source-text knowledge.
}

\begin{table*}[h]
\centering\small
\caption{{Title Repetition Rate: Urdu vs.\ English.}}
\label{tab:english_trr}
\begin{tabular}{lccccc}
\hline
\textbf{{Model}} & \textbf{{Ur.\ TRR}} & \textbf{{En.\ TRR (full)}} & \textbf{{En.\ TRR (2k w)}} & \textbf{{Ur.\ Cnt}} & \textbf{{En.\ Cnt}} \\
\hline
{GPT      }& {0.40 }& {0.069 }& {0.049 }& {6.06 }& {4.00 }\\
{QWEN     }& {0.40 }& {0.110 }& {0.070 }& {5.19 }& {2.86 }\\
{DeepSeek }& {0.43 }& {0.124 }& {0.105 }& {8.12 }& {4.00 }\\
\hline
\textbf{{Overall}} & \textbf{{0.41}} & \textbf{{0.101}} & \textbf{{0.075}} & \textbf{{6.46}} & \textbf{{3.62}} \\
\hline
\end{tabular}
\end{table*}

{Urdu TRR is 3.5$\times$ to 5.8$\times$ higher than English TRR
across all three models, aligning directly with the Lexical
Title Unpacking finding. Raw counts confirm this independently of
story length (Urdu mean 6.46 vs.\ English mean 3.62), and the
pattern holds after truncating English stories to 2}{{,}}{000 words,
so it is not an artefact of story-length variation.
}

\begin{table*}[h]
\centering\small
\caption{{Protagonist gender: Urdu vs.\ English.}}
\label{tab:english_gender}
\begin{tabular}{lcccccc}
\hline
\textbf{{Model}} & \textbf{{Ur.\ M}} & \textbf{{Ur.\ F}} & \textbf{{Ur.\ M:F}} & \textbf{{En.\ M}} & \textbf{{En.\ F}} & \textbf{{En.\ M:F}} \\
\hline
{GPT      }& {22 }& {7 }& {3.1:1 }& {2 }& {5 }& {0.4:1 }\\
{QWEN     }& {26 }& {5 }& {5.2:1 }& {3 }& {4 }& {0.75:1 }\\
{DeepSeek }& {22 }& {7 }& {3.1:1 }& {1 }& {6 }& {0.17:1 }\\
\hline
\textbf{{Overall}} & \textbf{{70}} & \textbf{{19}} & \textbf{{3.7:1}} & \textbf{{6}} & \textbf{{15}} & \textbf{{0.4:1}} \\
\hline
\end{tabular}
\end{table*}

{The Urdu male-to-female ratio (3.7:1) reverses to 0.4:1 in
English, suggesting the male-default pattern in Urdu is
language-conditioned rather than universal. We note a
title-selection confound: several TSST titles have canonical
female protagonists (}\emph{{The Yellow Wallpaper}}{, }\emph{{The Story
of an Hour}}{, }\emph{{Boys and Girls}}{), which partially accounts for
the English female majority; the comparison should therefore be
read as indicative rather than a clean cross-lingual experiment.
On name diversity, QWEN used ``Ayesha'' in 5 of 7 English
stories and DeepSeek used ``Zara'' in 4 of 7: despite the
language change, models continue to draw from a limited name
inventory.
}

\paragraph{{Pakistani vocabulary in English stories.}} {The
generation prompt specified a Pakistani cultural context. In the
English stories, models signalled this setting through Urdu
loanwords embedded in English prose, including ``haveli'',
``mohalla'', ``rishta'', ``nihari'', ``Allah ka shukar'', and a
dupatta's }\emph{{pallu}}{. This is the inverse of the pattern
documented in Urdu (Section~\ref{sec:cultural}), where Pakistani
surface vocabulary was present but cultural grounding remained
shallow: when writing in English about a Pakistani setting,
models appear to rely on lexical markers as a proxy for cultural
knowledge rather than producing contextually situated narrative
content.
}

\section{{Cross-Family Judge Validation: Full Results}}
\label{app:cross_judge}

{To address same-family-judge concerns, the memorisation-scoring
and character-network-extraction pipelines were re-run with
Gemini-3.5-Flash as an out-of-family judge on all 36 story pairs
and all 93 stories respectively.
}

\begin{table}[h]
\centering\small
\caption{{Memorisation results: GPT-5.4 vs.\ Gemini-3.5-Flash.}}
\label{tab:cross_judge_mem}
\begin{tabular}{lcc}
\hline
\textbf{{Metric}} & \textbf{{GPT-5.4}} & \textbf{{Gemini}} \\
\hline
{GPT faithful retelling (\%)      }& {25.0  }& {25.0  }\\
{DeepSeek faithful retelling (\%) }& {25.0  }& {25.0  }\\
{QWEN faithful retelling (\%)     }& \phantom{0}{8.3 }& \phantom{0}{8.3 }\\
{GPT composite score              }& {0.500 }& {0.364 }\\
{DeepSeek composite score         }& {0.430 }& {0.347 }\\
{QWEN composite score             }& {0.270 }& {0.162 }\\
{Cross overlap (GPT--DeepSeek)    }& {0.450 }& {0.388 }\\
{Cross overlap (GPT--QWEN)        }& {0.330 }& {0.217 }\\
{Cross overlap (DeepSeek--QWEN)   }& {0.340 }& {0.271 }\\
\hline
\end{tabular}
\end{table}

{Faithful-retelling percentages are identical across both judges
for all three models. Composite scores and cross-model overlap
are somewhat lower under Gemini, but direction and model
ordering are preserved (GPT~$>$~DeepSeek~$>$~QWEN). Categorical
memorisation findings are therefore robust to judge identity,
while continuous scores should be read as approximate.
}

\begin{table}[h]
\centering\small
\caption{{Character-network metrics: GPT-5.4 vs.\ Gemini-3.5-Flash.}}
\label{tab:cross_judge_network}
\begin{tabular}{lcc}
\hline
\textbf{{Metric}} & \textbf{{GPT-5.4}} & \textbf{{Gemini}} \\
\hline
{GPT positivity ratio       }& {0.780 }& {0.835 }\\
{DeepSeek positivity ratio  }& {0.820 }& {0.806 }\\
{QWEN positivity ratio      }& {0.820 }& {0.877 }\\
{GPT mean signed weight     }& {0.290 }& {0.442 }\\
{DeepSeek mean signed weight}& {0.370 }& {0.458 }\\
{QWEN mean signed weight    }& {0.380 }& {0.572 }\\
{Jaccard (GPT--DeepSeek)    }& {0.013 }& {0.008 }\\
{Jaccard (GPT--QWEN)        }& {0.018 }& {0.022 }\\
{Jaccard (DeepSeek--QWEN)   }& {0.013 }& {0.012 }\\
\hline
\end{tabular}
\end{table}

{Positivity ratios under Gemini are consistent with the GPT-5.4
results in both direction and model ordering, and all values
remain well above 0.80. Cross-model Jaccard overlap remains near
zero under both judges (median 0.0 across all pairs), confirming
that the character-diversity finding is robust to judge identity.
Mean signed edge weight is somewhat higher under Gemini,
indicating that intensity estimates are more sensitive to judge
choice than directional findings are.
}

\paragraph{{GPT-story error detection.}} {To cross-validate
linguistic/semantic error detection against an out-of-family
judge, Gemini-3.5-Flash was run on the 31 GPT-generated stories.
Gemini flagged 176 errors, of which 149 were confirmed correct by
a native Urdu speaker on manual review (84.7\% precision); the
remaining 27 were false positives arising from legitimate
spelling variants and acceptable space-deletion patterns in Urdu
orthography. This validation was conducted only for GPT-generated
stories due to budget constraints (see Limitations); the
agreement between Gemini and human annotation nonetheless
provides evidence that the error patterns identified in
Section~\ref{sec:networks} are not artefacts of same-family
judge preference.
}

\section{{Human Validation of Memorisation Labels}}
\label{app:human_validation}

{A native Urdu speaker manually reviewed 13 story pairs, covering
all three memorisation-label categories (PR, PI, OR) and all four
title categories, and compared their own judgement against the
GPT-4o-mini judge's categorical label.
}

\begin{table}[h]
\centering\scriptsize
\caption{{Human validation of memorisation labels (13 pairs).
PR = Plot Reproduction, PI = Partial Inheritance, OR = Original.}}
\label{tab:human_validation}
\begin{tabular}{clllccc}
\hline
\textbf{{\#}} & \textbf{{ID}} & \textbf{{Model}} & \textbf{{Cat.}} & \textbf{{Judge}} & \textbf{{Human}} & \textbf{{Match}} \\
\hline
{1  }& {13 }& {GPT      }& {Fable }& {PR }& {PR }& {Yes }\\
{2  }& {13 }& {QWEN     }& {Fable }& {PR }& {PR }& {Yes }\\
{3  }& {14 }& {DeepSeek }& {Fable }& {PR }& {PR }& {Yes }\\
{4  }& {22 }& {GPT      }& {Fable }& {PR }& {PR }& {Yes }\\
{5  }& {25 }& {DeepSeek }& {TSST  }& {PR }& {PR }& {Yes }\\
{6  }& {9  }& {GPT      }& {TSST  }& {OR }& {OR }& {Yes }\\
{7  }& {11 }& {GPT      }& {UST   }& {OR }& {OR }& {Yes }\\
{8  }& {17 }& {DeepSeek }& {UST   }& {OR }& {OR }& {Yes }\\
{9  }& {20 }& {GPT      }& {UST   }& {PI }& {PI }& {Yes }\\
{10 }& {21 }& {DeepSeek }& {UST   }& {PI }& {PI }& {Yes }\\
{11 }& {25 }& {QWEN     }& {TSST  }& {PI }& {OR }& {No }\\
{12 }& {31 }& {DeepSeek }& {UST   }& {PR }& {OR }& {No }\\
{13 }& {17 }& {GPT      }& {UST   }& {PI }& {OR }& {No }\\
\hline
\end{tabular}
\end{table}

{Agreement is 10/13 (76.9\%). All four Fable cases and all OR
cases assigned by the judge agree exactly with human judgement
(3/3). All three mismatches occur in UST titles, where the judge
over-estimates plot reproduction (assigning PR or PI where the
human judged OR), consistent with the Section~\ref{sec:tractability}
observation that the automatic judge is most prone to error on
titles it has never seen a source text for.
}

\section{{Extended Examples: Context Mismatch and Semantic Errors}}
\label{app:extended_examples}

{This appendix collects additional illustrative examples for the
error sub-categories summarised in
Section~\ref{sec:context_mismatch}, beyond the one or two
examples given in the main text for space.
}

\paragraph{{Kinship-term misuse (additional examples).}} {A
DeepSeek story introduces a maternal-side visit but uses
}\foreignlanguage{arabic}{دادی اماں} {(}\textit{{dādī ammā}}{,
``paternal grandmother'') for the grandmother encountered there.
QWEN uses }\foreignlanguage{arabic}{ابا جان} {(}\textit{{abbā jān}}{,
``father, respectful'') for an elderly stranger across multiple
stories, an over-generalised address form.
}

\paragraph{{Long-range gender breakdown (additional example).}} {In
one DeepSeek story, two figures including a girl are jointly
referred to as }\foreignlanguage{arabic}{دو نوجوان} {(}\textit{{do
nau-jawān}}{, ``two young men''), a masculine plural noun applied
across a mixed-gender pair.
}

\paragraph{{Cultural and religious anachronism (additional
examples).}} {Another QWEN story describes a moon-sighting
occurring in the morning, inconsistent with the lunar-sighting
practice it is meant to depict. A GPT story promotes a chef to
}\foreignlanguage{arabic}{کیپٹن} {(}\textit{{kepṭan}}{, ``captain''), a
term reserved for restaurant front-of-house staff in Pakistani
usage rather than kitchen roles.
}
\section{Space Deletion Errors}
\label{app:space_deletion}
GPT fuses
\foreignlanguage{arabic}{ہو} (\textit{ho}, ``be'')
constructions at substantial rates
(\foreignlanguage{arabic}{ہوگیا} 41\%,
\foreignlanguage{arabic}{ہوجاتا} 40\%,
\foreignlanguage{arabic}{ہوگا} 67\%), while DeepSeek
and QWEN write the same forms with a space in nearly
every instance. The pattern inverts for
\foreignlanguage{arabic}{آ} (\textit{ā}, ``come'')
auxiliaries: DeepSeek fuses
\foreignlanguage{arabic}{آگیا} 44\% and
\foreignlanguage{arabic}{آگئے} 67\% of the time, while
GPT and QWEN remain near zero. QWEN shows almost no
fusion of either family. Each model therefore
mishandles a different closed inventory of compounds
while writing the rest correctly, indicating that the
errors reflect sub-word vocabulary choices in each
model's tokeniser rather than a shared weakness in
Urdu generation. Crucially, this distinction matters
downstream: whitespace-based Urdu NLP pipelines treat
\foreignlanguage{arabic}{ہوگیا} (fused) and
\foreignlanguage{arabic}{ہو گیا} (spaced) as different
tokens, so if LLM output is reused as synthetic
training data, a model's fusion preferences gain
additional statistical support and risk
self-reinforcing contamination of the language's
digital record.

\section{Tractability: GPT-5.5 Remediation Prompt}
\label{app:tractability_prompt}

The post-editing pass described in
Section~\ref{sec:tractability} used the system prompt
reproduced below. The model was issued one call per
story with a JSON list of numbered sentences; the
response was constrained to a JSON object with a single
\texttt{results} array, with one entry per detected issue.
The prompt enforces minimal edits (a single inserted
space or a single replaced word) and includes a self-check
for \texttt{SpaceDeletion} to suppress punctuation-adjacent
false positives. Sentence splitting, JSON parsing, and a
post-hoc validator that rejected any
\texttt{SpaceDeletion} whose inserted space did not occur
between two Urdu letters were applied around this prompt.

\begin{quote}
\small

You are an expert Urdu linguist and proofreader.\\
You will receive a JSON list of numbered Urdu sentences
from one story. Detect the following issue types and
fix them with MINIMAL changes.

\medskip
Error Types

\medskip
\textbf{1. SpaceDeletion} (a.k.a.\ Word Segmentation
error). Two adjacent Urdu WORDS are written as a single
token because the space BETWEEN THE TWO WORDS is
missing. This is ONLY about word-to-word boundaries.

\smallskip
Flag examples (word-word fusion):
\begin{itemize}
  \item \foreignlanguage{arabic}{چاہتیہوں}
        $\rightarrow$
        \foreignlanguage{arabic}{چاہتی ہوں}

  \item \foreignlanguage{arabic}{گھرگیا}
        $\rightarrow$
        \foreignlanguage{arabic}{گھر گیا}

  \item \foreignlanguage{arabic}{ہوگیاہے}
        $\rightarrow$
        \foreignlanguage{arabic}{ہو گیا ہے}

  \item \foreignlanguage{arabic}{میںنے}
        $\rightarrow$
        \foreignlanguage{arabic}{میں نے}

  \item \foreignlanguage{arabic}{خوبصورتلڑکی}
        $\rightarrow$
        \foreignlanguage{arabic}{خوبصورت لڑکی}
\end{itemize}

Do NOT flag any of these:
\begin{itemize}
  \item Missing space AROUND punctuation
        (\foreignlanguage{arabic}{،}\
        \foreignlanguage{arabic}{۔}\ ?\ !\ .\ :)
        e.g.\
        \foreignlanguage{arabic}{ہوگا،ہوچکا}
        is NOT a \texttt{SpaceDeletion} issue.

  \item Missing space around quotes
        (\verb|"|, ”, “, ', ’, ‘),
        brackets, parentheses, dashes, or numbers.

  \item Missing line breaks, paragraph breaks, or
        dialogue formatting.

  \item Long or multi-clause sentences.
\end{itemize}

Fix rule: insert ONLY the single missing space between
the two fused WORDS. Do not touch any punctuation,
quotes, or formatting. \texttt{correct\_sentence} must
differ from \texttt{incorrect\_sentence} by exactly one
(or a few) inserted SPACE characters between word
characters.

\medskip
\textbf{2. SpellingIssue.} A single word is misspelled.

\smallskip
\foreignlanguage{arabic}{ہیریو}
$\rightarrow$
\foreignlanguage{arabic}{ہیرو};
\qquad
\foreignlanguage{arabic}{مسئیبت}
$\rightarrow$
\foreignlanguage{arabic}{مصیبت}.

\smallskip
Fix: change ONLY the misspelled word.

\medskip
\textbf{3. GrammarIssue.} Wrong case marking,
missing or wrong
\foreignlanguage{arabic}{نے}/%
\foreignlanguage{arabic}{کو}/%
\foreignlanguage{arabic}{سے},
verb agreement, etc.

\smallskip
\foreignlanguage{arabic}{اُنہوں نے مجھے گرمجوشی سے خیرمقدم کیا}
$\rightarrow$
\foreignlanguage{arabic}{اُنہوں نے میرا گرمجوشی سے خیرمقدم کیا}.

\smallskip
Fix: minimal grammatical correction.

\medskip
\textbf{4. Agreement Error.} Gender, number, or
oblique-case agreement error between a noun and its
modifier (adjective, demonstrative, quantifier, or
possessive).

\smallskip
\foreignlanguage{arabic}{ایک چھوٹی سی لینگر خانہ}
$\rightarrow$
\foreignlanguage{arabic}{ایک چھوٹا سا لنگر خانہ};

\foreignlanguage{arabic}{اتنی خراب کر دیتی ہے}
$\rightarrow$
\foreignlanguage{arabic}{اتنا خراب کر دیتی ہے}.

\medskip
\textbf{5. Semantic Anomaly.} Sentence is grammatically
OK but a word choice makes it meaningless or wrong in
context.

\smallskip
\foreignlanguage{arabic}{آواز میں ایک عجیب سی کانپ تھی}
$\rightarrow$
\foreignlanguage{arabic}{آواز میں ایک عجیب سی کپکپاہٹ تھی}.

\smallskip
Fix: replace ONLY the offending word.

\medskip
{======= GENERAL RULES =====}

\begin{itemize}
  \item One sentence may contain MULTIPLE issues
        $\rightarrow$ emit one entry per issue.

  \item If a sentence has NO issue, emit nothing for
        it.

  \item \texttt{incorrect\_sentence} MUST equal the
        original sentence text exactly.

  \item \texttt{correct\_sentence} MUST contain ONLY
        the minimal fix for THAT issue.

  \item Never split or merge sentences. Never reformat
        dialogue or quotes.

  \item Self-check before emitting
        \texttt{SpaceDeletion}: are the two characters
        around the inserted space BOTH Urdu letters,
        and was there NO whitespace and NO punctuation
        between them in the original? If NO, do NOT
        emit \texttt{SpaceDeletion}.
\end{itemize}

Respond ONLY with valid JSON, no markdown, no commentary:

\begin{verbatim}
{
  "results": [
    {
      "sentence_id": <int>,
      "error_tag": "<SpaceDeletion|SemanticAnomaly|SpellingIssue|GrammarIssue|AgreementError>",
      "incorrect_sentence": "<original sentence>",
      "correct_sentence": "<sentence with ONLY the minimal fix>"
    }
  ]
}
\end{verbatim}

\end{quote}
```

 end\end{document}